\pdfoutput=1
\documentclass[11pt]{article}

\usepackage[final]{acl}

\usepackage{times}    
\usepackage{helvet}   
\usepackage{courier}  
\usepackage{latexsym}
\usepackage[T1]{fontenc}
\usepackage[utf8]{inputenc}
\usepackage{microtype}
\usepackage{inconsolata}

\usepackage{amsmath}
\usepackage{amssymb}
\usepackage{booktabs}
\usepackage{tabularx}
\usepackage{array}
\usepackage{multirow}
\usepackage{enumitem}

\newcolumntype{Y}{>{\centering\arraybackslash}X}

\usepackage{graphicx}
\usepackage{float}
\usepackage{caption} 
\usepackage{subcaption}
\usepackage{newfloat}
\usepackage{orcidlink}
\usepackage{xcolor}
\usepackage[most]{tcolorbox}
\usepackage{ragged2e}

\usepackage{url}
\usepackage{natbib} 

\definecolor{PromptBlack}{HTML}{111827}
\definecolor{PromptBlue}{HTML}{1D4ED8}
\definecolor{PromptPurple}{HTML}{7C3AED}
\definecolor{PromptTeal}{HTML}{047857}
\definecolor{PromptOrange}{HTML}{B45309}
\definecolor{PromptGray}{HTML}{6B7280}
\definecolor{PromptLine}{HTML}{D1D5DB}
\definecolor{PromptBg}{HTML}{FAFAFA}

\newtcolorbox{promptbox}[2][]{%
    enhanced,
    breakable,
    width=\linewidth,
    colback=PromptBg,
    colframe=PromptBlack,
    colbacktitle=PromptBlack,
    coltitle=white,
    fonttitle=\bfseries\footnotesize,
    fontupper=\ttfamily\fontsize{6.6pt}{7.4pt}\selectfont,
    before upper={\RaggedRight\sloppy},
    title={#2},
    boxrule=0.65pt,
    arc=1.8mm,
    outer arc=1.8mm,
    left=5pt,
    right=5pt,
    top=4pt,
    bottom=4pt,
    toptitle=3pt,
    bottomtitle=3pt,
    before skip=5pt,
    after skip=5pt,
    halign title=center,
    #1
}

\providecommand{\pkw}[1]{}
\renewcommand{\pkw}[1]{\textcolor{PromptBlue}{\textbf{#1}}}

\providecommand{\pfld}[1]{}
\renewcommand{\pfld}[1]{\textcolor{PromptPurple}{\textbf{#1}}}

\providecommand{\pjson}[1]{}
\renewcommand{\pjson}[1]{%
    \textcolor{PromptTeal}{\ttfamily\fontsize{6.3pt}{7.1pt}\selectfont #1}%
}

\providecommand{\pwarn}[1]{}
\renewcommand{\pwarn}[1]{\textcolor{PromptOrange}{\textbf{\texttt{#1}}}}

\providecommand{\fewshot}[5]{}
\renewcommand{\fewshot}[5]{%
\noindent
\begin{minipage}[t]{0.055\linewidth}
\textcolor{PromptOrange}{\textbf{#1}}
\end{minipage}
\begin{minipage}[t]{0.925\linewidth}
\RaggedRight\sloppy
\textcolor{PromptBlue}{\textbf{In:}}\\[-0.4mm]
#2

\vspace{0.4mm}
\textcolor{PromptBlue}{\textbf{Out:}} #3
\end{minipage}

\vspace{0.7mm}
{\color{PromptLine}\hrule height 0.25pt}
\vspace{0.7mm}
}

\usepackage{tikz}
\usetikzlibrary{arrows.meta,positioning,shapes.geometric,fit,backgrounds,patterns}

\usepackage{pgfplots}
\pgfplotsset{compat=1.18}
\usepgfplotslibrary{groupplots}

\definecolor{cbBlue}{HTML}{1F77B4}
\definecolor{cbOrange}{HTML}{FF7F0E}
\definecolor{cbGreen}{HTML}{2CA02C}
\definecolor{cbRed}{HTML}{D62728}
\definecolor{cbPurple}{HTML}{9467BD}

\tikzset{
  small/.style={font=\scriptsize},
  gate/.style={
    rectangle, rounded corners=2pt,
    draw=cbBlue, thick, fill=cbBlue!6,
    align=left, inner sep=4pt,
    text width=.66\columnwidth
  },
  accept/.style={
    rectangle, rounded corners=2pt,
    draw=cbGreen, thick, fill=cbGreen!10,
    align=center, inner sep=4pt,
    minimum width=.80\columnwidth
  },
  reject/.style={
    rectangle, rounded corners=2pt,
    draw=cbRed, thick, fill=cbRed!10,
    align=center, inner sep=3pt,
    minimum width=.24\columnwidth
  },
  note/.style={
    rectangle, rounded corners=2pt,
    draw=cbOrange, thick, fill=cbOrange!8,
    align=left, inner sep=4pt,
    text width=.80\columnwidth
  },
  arrow/.style={-Stealth, very thick}
}

\usepackage{listings}
\DeclareCaptionStyle{ruled}{labelfont=normalfont,labelsep=colon,strut=off} 
\floatstyle{ruled}
\newfloat{listing}{tb}{lst}{}
\floatname{listing}{Listing}

\makeatletter
\@ifundefined{href}{%
  \def\href#1#2{#2}%
}{%
  \renewcommand{\href}[2]{#2}%
}
\@ifundefined{path}{}{}
\@ifundefined{autoref}{\def\autoref#1{Section~\ref{#1}}}{%
  \renewcommand{\autoref}[1]{Section~\ref{#1}}%
}
\@ifundefined{texorpdfstring}{\def\texorpdfstring#1#2{#1}}{%
  \renewcommand{\texorpdfstring}[2]{#1}%
}
\makeatother

\usepackage{algorithm}
\usepackage{algorithmic}

\title{C³T: Counterfactual Causal Reasoning for Sentiment Shifts in Social-Media Conversation Trees}

\author{
  S M Rafiuddin\,\orcidlink{0000-0001-9404-1556} \\
  Department of Computer Science \\
  Oklahoma State University \\
  Stillwater, OK 74078, USA \\
  \texttt{srafiud@okstate.edu}
  \And
  Atriya Sen\,\orcidlink{0000-0002-1640-5905} \\
  Department of Computer Science \\
  Oklahoma State University \\
  Stillwater, OK 74078, USA \\
  \texttt{atriya.sen@okstate.edu}
}

\begin{document}
\maketitle
\begin{abstract}
Sentiment in social-media threads does not only vary across posts; it \emph{shifts} as users react to \emph{claims}, \emph{corrections}, \emph{evidence}, and \emph{hostility} within a branching reply tree. We study \emph{why} sentiment changes in rumor-centric conversation trees by treating discourse moves (e.g., \texttt{denial/correction}, \texttt{evidence/link}, \texttt{toxicity/attack}) as candidate interventions and asking (i) what sentiment a reply expresses, (ii) whether the sentiment shifts relative to its parent, and (iii) which prior message most plausibly drove the reply's sentiment. To support this setting, we introduce \textsc{CaSiRe}, a causal sentiment reasoning layer over public rumor conversation datasets that adds post-level sentiment labels, induced parent--child shift labels, calibrated multi-label intervention tags, and explicitly annotated causal-source labels. We then propose C$^{3}$T (\textbf{C}ounterfactual \textbf{C}ausal \textbf{C}onversation \textbf{T}ransformer), a \textbf{thread-structured temporal model} that jointly predicts node sentiment and shifts, learns sparse ancestor attribution, and supports counterfactual queries by forcing conversational intervention embeddings on or off to estimate potential outcomes. Under an \emph{event-level split}, C$^{3}$T improves out-of-event robustness and attribution over text-only, graph-based, and temporal baselines, and yields interpretable model-based effects: denials/corrections and evidence reduce downstream negativity, while toxicity increases it. We also benchmark \emph{open-weight LLM prompting baselines} and find that added conversational context helps, but attribution remains less reliable, motivating \textbf{structure-aware counterfactual modeling} for social-media analysis.
\end{abstract}

\section{Introduction}
\label{sec:intro}

Online social-media conversations are \emph{interactional}: users react to one another and to unfolding events, and threads can converge or polarize in \emph{sentiment} and \emph{stance}. These dynamics matter because misinformation diffuses differently than accurate information on Twitter/X~\citep{VosoughiScience2018}, and exposure to others' emotional expressions can shift what people express~\citep{kramer2014emotional}. Thus, the key question is not only \emph{what} sentiment is expressed, but \emph{why} sentiment shifts within a reply thread, e.g., whether negativity is driven by a \emph{new claim}, a \emph{correction}, or a \emph{hostile exchange}.

Conversation trees provide a natural substrate: a source post induces a \emph{branching cascade} where local parent--child interactions can propagate globally. Rumour and claim-centric benchmarks capture such trees in breaking-news contexts~\citep{ZubiagaCOLING2016,RumourEval2019}, while prior work has largely focused on \emph{stance/veracity}, including detection, stance classification, and verification~\citep{zubiaga2018survey}. In contrast, the \emph{affective trajectory}, how sentiment changes over time and which conversational inputs plausibly drive that change, remains under-modeled.

Attributing sentiment change is fundamentally \emph{causal}, but observational social-media data contain endogenous participation and latent confounders. Recent work has formalized causal questions in online interactions, including effects of tone and conversational behaviors~\citep{SridharGetoorIJCAI2019,ZhangCSCW2020}. However, most formulations operate at coarse conversational attributes rather than discourse-move-specific effects within evolving reply trees. Related work is discussed in Appendix~\ref{sec:related}.

We propose a framework for causal sentiment reasoning that treats conversational discourse moves as candidate interventions and estimates their downstream impact on expressed sentiment. We represent threads as \emph{time-stamped trees}, learn structure- and time-aware representations, and predict node-level sentiment together with interpretable estimates of which prior conversational moves plausibly shifted sentiment. We evaluate on public rumour benchmarks~\citep{ZubiagaCOLING2016,RumourEval2019} and compare against neural and open-weight LLM prompting baselines, including Llama~3~\citep{Llama3Meta2024}, Mistral~7B~\citep{JiangMistral7B2023}, and Qwen~2.5~\citep{yang2024qwen25}.

\paragraph{Contributions.}
(1) We formulate \emph{causal sentiment reasoning in conversation trees} as a conversational-intervention problem grounded in public rumour datasets.
(2) We propose a \textbf{thread-structured temporal model} with explicit forced-intervention counterfactual estimation over conversational intervention types.
(3) We compare against \emph{text-only}, \emph{graph-based}, \emph{temporal}, and \emph{open-LLM prompting} baselines, and provide ablations isolating counterfactual loss, intervention tags, event-level invariance, and ancestor-window attribution.

\section{Tasks and Causal Setup}
\label{sec:setup}

\subsection{Conversation tree and notation}
\label{subsec:tree_def}
We model each social-media discussion thread as a \textbf{rooted, time-stamped conversation tree} $\mathbf{G}=(\mathbf{V},\mathbf{E})$, where each node $i\in \mathbf{V}$ corresponds to a \emph{post/comment} and each directed edge $(p\rightarrow i)\in \mathbf{E}$ indicates that post $i$ replies to its parent $p=\pi(i)$. Each node $i$ has timestamp $\tau_i$ (posting time) and observed covariates $\mathbf{X}_i$ (e.g., \texttt{text}, \texttt{author/platform metadata} when available, \texttt{depth/position features}, and \texttt{contextual summaries}). We denote the ancestor set of node $i$ by $\mathcal{A}(i)$ and its depth by $d(i)$. For causal estimands defined over \emph{downstream} effects, we use the $k$-hop descendants of $i$, defined as $\mathcal{D}_k(i)=\{j\in\mathbf{V}:\mathrm{dist}(i,j)\le k \wedge i\in\mathcal{A}(j)\}$, where $\mathrm{dist}(\cdot,\cdot)$ is the shortest-path length along directed reply edges.

\subsection{Outcomes: sentiment and sentiment shift}
\label{subsec:outcomes}
We consider two observational prediction targets.
First, each node $i$ has a three-class sentiment label $Y_i \in \{\texttt{negative}, \texttt{neutral}, \texttt{positive}\}$.
Second, each reply edge $(p\rightarrow i)$ has a \emph{sentiment-shift} label
$\mathbf{S}_{p\rightarrow i} \in \{-1,0,+1\}$ capturing change from parent to child,
representing \emph{down}, \emph{no-change}, and \emph{up} sentiment shifts, respectively.
We induce $S_{p\to i}$ by comparing the parent and child sentiment labels using the natural ordering \texttt{negative} $<$ \texttt{neutral} $<$ \texttt{positive}.

\subsection{Treatments: conversational interventions}
\label{subsec:treatments}
Beyond predicting $Y_i$ and $S_{p\rightarrow i}$, we aim to reason about how specific
\emph{conversational moves} affect downstream sentiment. We represent each post $i$ with a \textbf{multi-label
intervention (treatment) vector} $T_i \in \{0,1\}^M$, where each component indicates the presence
of an intervention type (\emph{multi-hot encoding}). In our primary taxonomy, $M$ includes:
(i) \texttt{claim/rumor assertion}, (ii) \texttt{denial/correction}, (iii) \texttt{evidence/link provision},
(iv) \texttt{authority citation}, (v) \texttt{toxicity/attack}, (vi) \texttt{sarcasm/irony}, (vii) \texttt{question/challenge},
and (viii) \texttt{derail/off-topic move}. A node may carry \emph{multiple interventions} (e.g., a correction
with an external evidence link). For causal estimands below, we study the effect of toggling a
single intervention type $m$ while holding other components fixed.

\subsection{Causal estimands}
\label{subsec:estimands}
We adopt the \textbf{potential-outcomes framework} \citep{Rubin1974,RosenbaumRubin1983}. Let $Y_j(\mathbf{t})$ denote the possibly \emph{counterfactual sentiment outcome} for node $j$ under a hypothetical intervention assignment $\mathbf{t}$. Since we focus on the causal impact of intervention $m$ at a \textbf{source node} $i$ on \emph{downstream} sentiment expressed by its descendants, we define a descendant-level aggregated outcome. Let $\mathbb{I}[\cdot]$ be an indicator and define negativity as $\mathbb{I}[Y_j=\texttt{negative}]$. For a fixed hop budget $k$, the \textbf{average downstream negativity} from node $i$ is $\bar{Y}^{\mathrm{neg}}_{i}(k)=\frac{1}{|\mathcal{D}_k(i)|}\sum_{j\in\mathcal{D}_k(i)}\mathbb{I}[Y_j=\texttt{negative}]$, with the convention $\bar{Y}^{\mathrm{neg}}_{i}(k)=0$ if $\mathcal{D}_k(i)=\emptyset$. To isolate the effect of intervention type $m$, we consider two counterfactual assignments for the source node $i$: $T_i^{(m=1)}$ (\emph{force intervention $m$ present}) and $T_i^{(m=0)}$ (\emph{force it absent}), while keeping other components unchanged. The \textbf{$k$-hop average treatment effect (ATE)} for intervention $m$ is $\mathrm{ATE}_m(k)=\mathbb{E}\!\left[\bar{Y}^{\mathrm{neg}}_{i}(k)\!\left(T_i^{(m=1)}\right)-\bar{Y}^{\mathrm{neg}}_{i}(k)\!\left(T_i^{(m=0)}\right)\right]$, where the expectation is taken over source nodes and threads/events under the evaluation split. In this work, we restrict causal-effect reporting to downstream negativity as the primary descendant-level outcome. Conditional or heterogeneous effects by event type, platform, veracity, or conversational stage are treated as exploratory analyses and are not used as primary evidence unless explicitly reported with confidence intervals.

\subsection{Assumptions and scope}
\label{subsec:assumptions}
Our observational causal analysis assumes standard conditions \citep{Rubin1974,RosenbaumRubin1983}: (i) \textbf{Temporal precedence:} interventions at $i$ precede outcomes for descendants $j\in\mathcal{D}_k(i)$ (\texttt{timestamps/reply edges}). (ii) \textbf{Consistency:} observed outcomes equal the corresponding potential outcomes under the realized assignment. (iii) \textbf{Positivity:} within covariate strata used, intervention $m$ can occur and not occur with non-zero probability. (iv) \textbf{Limited interference:} effects are modeled primarily within $k$-hop descendants; cross-thread spillovers are treated as a limitation. (v) \textbf{Conditional ignorability (scope-limited):} conditioning on observed thread context (\texttt{text}, \texttt{position}, \texttt{time}, \texttt{history summaries}), assignment of $m$ at $i$ is as-if random for the downstream outcomes; we emphasize patterns stable across events/platforms.

\section[Method: C3T]{Method: \texorpdfstring{C$^{3}$T}{C3T}}

\label{sec:method}
We propose \textbf{C$^{3}$T} (\textbf{C}ounterfactual \textbf{C}ausal \textbf{C}onversation \textbf{T}ransformer), a \textbf{thread-structured model} that (i) predicts \emph{node sentiment} and \emph{edge-level sentiment shifts}, (ii) attributes each reply's sentiment to a small set of prior messages (\emph{ancestors}), and (iii) estimates \emph{counterfactual potential outcomes} under conversational interventions.

\subsection{Overview}
\label{subsec:method_overview}

Given a \textbf{conversation tree} $G=(V,E)$, each node $i\in V$ has \textbf{text} $x_i$, \textbf{timestamp} $\tau_i$, \textbf{structural position features} (e.g., \texttt{depth} $d(i)$ and \texttt{sibling index}), and optional \textbf{author/platform features} $\mathbf{a}_i$. We do \textbf{not} use gold or predicted parent sentiment labels as input features. Instead, \textbf{parent information} is incorporated through the \textbf{parent node representation} and the \textbf{reply-tree structure}, ensuring that the model receives the same observable information at \textbf{training} and \textbf{inference} time. C$^{3}$T outputs: (i) \textbf{\emph{sentiment logits}} $\hat{\mathbf{y}}_i$ for each node label $Y_i$, (ii) \textbf{\emph{shift logits}} $\hat{\mathbf{s}}_{p\rightarrow i}$ for each reply edge $(p\rightarrow i)$ and its shift label $S_{p\rightarrow i}$, (iii) an \textbf{ancestor attribution distribution} $\boldsymbol{\alpha}_i$ over candidate ancestors for reply node $i$, and (iv) \textbf{descendant-level counterfactual predictions} under \textbf{forced treatment-on} and \textbf{treatment-off} assignments, $\hat{\mathbf{y}}_j(T_i^{(m=1)})$ and $\hat{\mathbf{y}}_j(T_i^{(m=0)})$, for descendants $j\in\mathcal{D}_k(i)$ of a source node $i$ and intervention type $m$.

\begin{figure*}[t]
    \centering
    \includegraphics[width=0.88\linewidth]{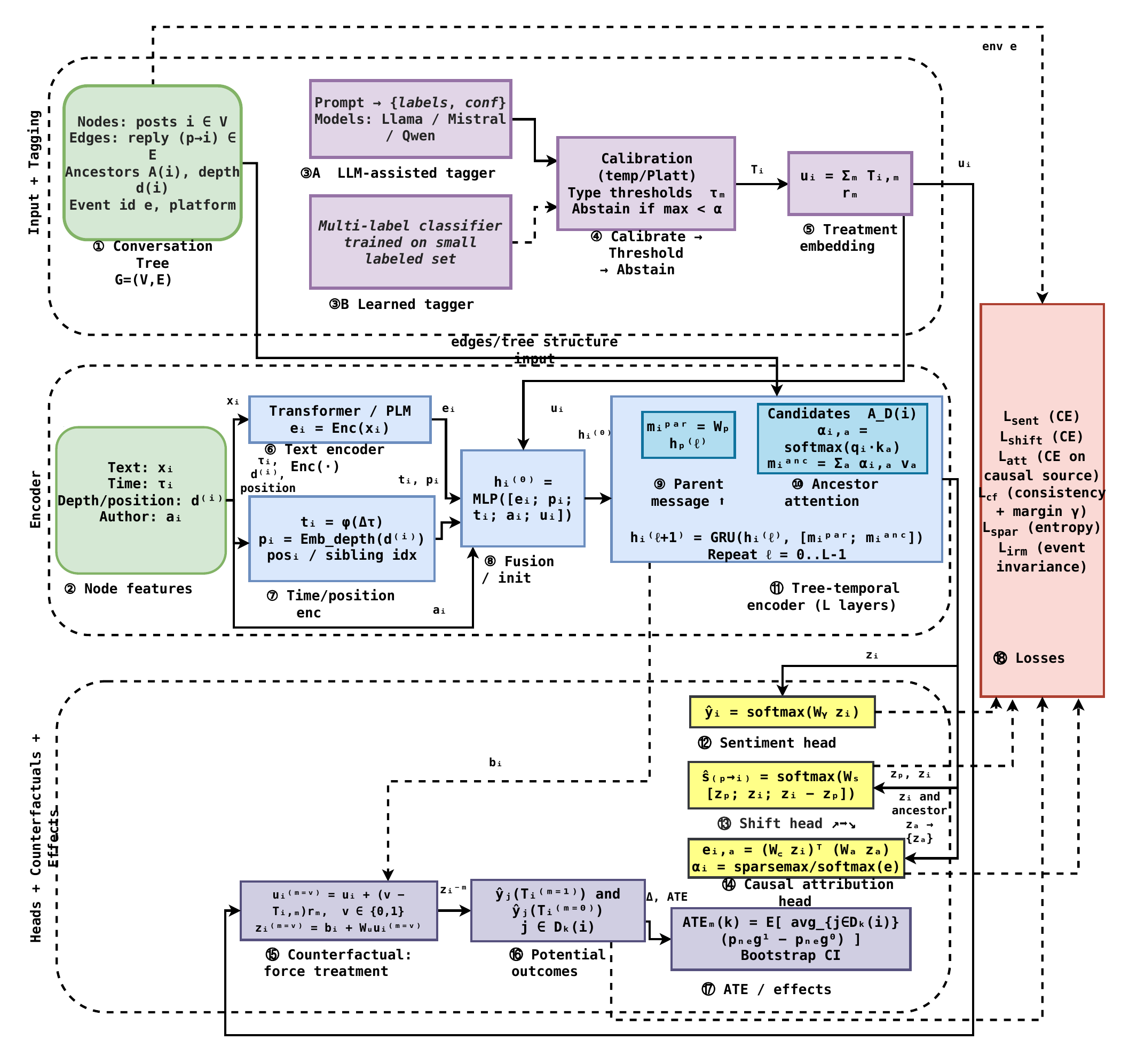}
    \vspace{-2mm}
\caption{\textbf{C$^{3}$T architecture.} A \emph{treatment tagger} yields the multi-label intervention vector $T_i$. A \emph{temporal/tree encoder} produces node representations using \textbf{parent message passing} and \textbf{ancestor attention}. Prediction heads estimate \emph{sentiment}, \emph{shift}, and \emph{sparse ancestor attribution}. Finally, forced on/off intervention embeddings yield potential-outcome predictions for downstream effect estimation.}
    \label{fig:c3t_arch}
    \vspace{-3mm}
\end{figure*}

\subsection{Intervention tagging module}
\label{subsec:method_tagging}
Each post $i$ receives a multi-label treatment vector $T_i\in\{0,1\}^M$ indicating intervention types (claim, denial/correction, evidence/link, authority cite, toxicity/attack, sarcasm, question/challenge, derail). We consider two tagging variants. \textbf{LLM-assisted tagging.} We use \texttt{Llama~3-8B} to output intervention labels with confidence scores \citep{Llama3Meta2024}. Let raw scores be $\tilde{s}_{i,m}\in[0,1]$. Because LLM confidences can be poorly calibrated, we calibrate scores using post-hoc scaling on a validation subset (temperature scaling \citep{GuoCalibration2017} or sigmoid/Platt scaling \citep{Platt1999}). After calibration we obtain $s_{i,m}$ and set $T_{i,m}=\mathbb{I}[s_{i,m}\ge \tau_m]$, where $\tau_m$ is type-specific and chosen on validation to optimize F1/coverage tradeoff. We additionally allow an \emph{abstain} option: if $\max_m s_{i,m}<\alpha$, we mark $T_i$ as unknown and exclude it from the counterfactual loss, following selective prediction principles \citep{GeifmanElYaniv2017}. \textbf{Learned tagger (ablation / alternative).} We train a lightweight multi-label classifier on a small manually labeled subset $\mathcal{L}$: $\mathbf{s}_i=\sigma(W_t\,\mathrm{Enc}(x_i)+\mathbf{b}_t)$, where $\mathrm{Enc}(\cdot)$ is the text encoder and $\sigma$ is elementwise sigmoid. This tagger replaces the LLM module and isolates the value of LLM-assisted weak supervision.

\subsection{Conversation encoder (structure + time)}
\label{subsec:method_encoder}
We encode each post with a \textbf{Transformer text encoder}~\citep{Vaswani2017}, producing $\mathbf{e}_i=\mathrm{Enc}(x_i)\in\mathbb{R}^d$, and add \emph{structural} and \emph{temporal} features $\mathbf{p}_i=\mathrm{Emb}_{\mathrm{depth}}(d(i))$ and $\mathbf{t}_i=\phi(\Delta\tau_i)$, where $\Delta\tau_i=\tau_i-\tau_{\pi(i)}$ for non-root replies and $\Delta\tau_i=0$ for root nodes. Here, $\phi(\cdot)$ is a continuous time encoding (e.g., \texttt{sinusoidal / MLP time features} as in temporal graph attention~\citep{XuTGAT2020}), $\mathrm{Emb}_{\mathrm{depth}}$ is a learned depth embedding, and $\mathbf{a}_i$ denotes optional author/platform features. We map the treatment vector $T_i$ to an embedding $\mathbf{u}_i=\sum_{m=1}^{M}T_{i,m}\mathbf{r}_m$, with learned $\mathbf{r}_m\in\mathbb{R}^d$, and initialize each node as $\mathbf{h}_i^{(0)}=\mathrm{MLP}([\mathbf{e}_i;\mathbf{p}_i;\mathbf{t}_i;\mathbf{a}_i;\mathbf{u}_i])$. No gold or predicted parent sentiment label is used as an input feature. We then perform $L$ layers of \textbf{tree-temporal updates} using parent message passing and ancestor attention. For non-root replies, the parent message is $\mathbf{m}^{\mathrm{par}}_i=\mathbf{W}_p\mathbf{h}_{\pi(i)}^{(\ell)}$, while for root nodes we set $\mathbf{m}^{\mathrm{par}}_i=\mathbf{0}$. Let $\mathcal{A}_D(i)$ be the $D$ nearest ancestors of $i$; if $\mathcal{A}_D(i)=\emptyset$, we set $\mathbf{m}^{\mathrm{anc}}_i=\mathbf{0}$, otherwise we compute $\alpha_{i,a}=\mathrm{softmax}_{a\in\mathcal{A}_D(i)}\left((\mathbf{W}_q\mathbf{h}_i^{(\ell)})^\top(\mathbf{W}_k\mathbf{h}_a^{(\ell)})/\sqrt{d}\right)$ and $\mathbf{m}^{\mathrm{anc}}_i=\sum_{a\in\mathcal{A}_D(i)}\alpha_{i,a}\mathbf{W}_v\mathbf{h}_a^{(\ell)}$. A sibling aggregation term can be added by pooling representations of nodes sharing the same parent. We update $\mathbf{h}_i^{(\ell+1)}=\mathrm{GRU}(\mathbf{h}_i^{(\ell)},[\mathbf{m}^{\mathrm{par}}_i;\mathbf{m}^{\mathrm{anc}}_i])$, following dynamic-graph design patterns that combine attention and recurrent updates for time-evolving neighborhoods~\citep{RossiTGN2020,XuTGAT2020}, and use $\mathbf{z}_i=\mathbf{h}_i^{(L)}$ as the final node representation.

\subsection{Prediction heads}
\label{subsec:method_heads}
\textbf{Sentiment head.} We predict the \emph{sentiment distribution} for node $i$ as $\hat{\mathbf{y}}_i = \mathrm{softmax}(\mathbf{W}_y \mathbf{z}_i + \mathbf{b}_y)$. \textbf{Shift head.} For edge $(p\rightarrow i)$, we predict the \emph{sentiment shift} using paired parent--child features as $\hat{\mathbf{s}}_{p\rightarrow i} = \mathrm{softmax}\!\left(\mathbf{W}_s [\mathbf{z}_p;\mathbf{z}_i;\mathbf{z}_i-\mathbf{z}_p] + \mathbf{b}_s\right)$.

\subsection{Counterfactual representation learning}
\label{subsec:method_counterfactual}

Our goal is to avoid purely correlational use of $T_i$ and instead learn representations that support \emph{counterfactual} queries: ``how would the sentiment prediction change if intervention $m$ were forced present or absent?'' For intervention type $m$, we define a forced assignment $T_i^{(m=v)}$ for $v\in\{0,1\}$ such that $T_{i,m}^{(m=v)}=v$ and $T_{i,r}^{(m=v)}=T_{i,r}$ for all $r\neq m$. The corresponding treatment embedding is $\mathbf{u}_i^{(m=v)}=\mathbf{u}_i+(v-T_{i,m})\mathbf{r}_m$, where $v=0$ gives the treatment-off counterfactual and $v=1$ gives the treatment-on counterfactual, including cases where the intervention was originally absent. To reduce compute, we use a \textbf{two-stream decomposition}, $\mathbf{z}_i^{(m=v)}=\mathbf{b}_i+\mathbf{W}_u\mathbf{u}_i^{(m=v)}$, where $\mathbf{b}_i$ is the \emph{structure+text representation} produced by the encoder when $T_i$ is omitted from the input; specifically, $\mathbf{b}_i$ is computed from $[\mathbf{e}_i;\mathbf{p}_i;\mathbf{t}_i;\mathbf{a}_i]$, after which the treatment embedding is injected additively. We then obtain potential-outcome predictions as $\hat{\mathbf{y}}_i(T_i^{(m=v)})=\mathrm{softmax}(\mathbf{W}_y\mathbf{z}_i^{(m=v)}+\mathbf{b}_y)$ for $v\in\{0,1\}$. This design is inspired by representation-learning approaches for counterfactual inference that separate nuisance variation from treatment-sensitive factors~\citep{Johansson2016Counterfactual,Shalit2017CFR}. \textbf{Losses.} We optimize the weighted objective $\mathcal{L}=\mathcal{L}_{\mathrm{sent}}+\lambda_{\mathrm{shift}}\mathcal{L}_{\mathrm{shift}}+\lambda_{\mathrm{att}}\mathcal{L}_{\mathrm{att}}+\lambda_{\mathrm{cf}}\mathcal{L}_{\mathrm{cf}}+\lambda_{\mathrm{spar}}\mathcal{L}_{\mathrm{spar}}+\lambda_{\mathrm{irm}}\mathcal{L}_{\mathrm{irm}}$, where $\mathcal{L}_{\mathrm{sent}}=\sum_{i\in V}\mathrm{CE}(Y_i,\hat{\mathbf{y}}_i)$ and $\mathcal{L}_{\mathrm{shift}}=\sum_{(p\rightarrow i)\in E}\mathrm{CE}(S_{p\rightarrow i},\hat{\mathbf{s}}_{p\rightarrow i})$. For \textbf{counterfactual consistency}, we define $\delta_{i,m}=\|\hat{\mathbf{y}}_i(T_i^{(m=1)})-\hat{\mathbf{y}}_i(T_i^{(m=0)})\|_2^2$ and use the per-node penalty $\ell_{i,m}^{\mathrm{cf}}=T_{i,m}\max(0,\gamma-\delta_{i,m})+(1-T_{i,m})\beta\delta_{i,m}$, where $\gamma>0$ encourages non-trivial sensitivity for observed interventions and $\beta$ controls sensitivity to interventions absent from the observed post. The total counterfactual loss is $\mathcal{L}_{\mathrm{cf}}=\sum_{i\in V}\sum_{m=1}^{M}\ell_{i,m}^{\mathrm{cf}}$.

\subsection{Event-level invariance for out-of-distribution robustness}
\label{subsec:method_invariance}
Threads are grouped by events, and we treat each event $e$ as an environment $\mathcal{E}_e$. To reduce reliance on event-specific predictive cues, we use an \textbf{IRM-style penalty}~\citep{ArjovskyIRM2019}. Let $\mathcal{L}_{\mathrm{pred}}=\mathcal{L}_{\mathrm{sent}}+\lambda_{\mathrm{shift}}\mathcal{L}_{\mathrm{shift}}$ denote the combined prediction loss within an environment. We define the penalty as $\mathcal{L}_{\mathrm{irm}}=\sum_e\left\|\nabla_w\mathcal{L}_{\mathrm{pred}}^{(e)}(w\cdot\Phi_\theta)\big|_{w=1}\right\|_2^2$, where $\Phi_\theta$ denotes the learned representation feeding the prediction heads. This penalty is used for \textbf{event-level predictive invariance}; it does not by itself identify causal intervention effects.

\subsection{Causal attribution head}
\label{subsec:method_attribution}
For each reply node $i$, we predict which ancestor most plausibly caused its sentiment or shift. We consider candidate ancestors $\mathcal{A}_D(i)$ and compute scores $e_{i,a}=(\mathbf{W}_c\mathbf{z}_i)^\top(\mathbf{W}_a\mathbf{z}_a)$ for $a\in\mathcal{A}_D(i)$, then normalize to obtain an attribution distribution $\boldsymbol{\alpha}_i$. We use either softmax or a sparse normalizer (e.g., sparsemax) to encourage peaked distributions \citep{MartinsSparsemax2016}: $\boldsymbol{\alpha}_i=\mathrm{Norm}(\{e_{i,a}\}_{a\in\mathcal{A}_D(i)})$. Nodes with empty ancestor sets are excluded from attribution training and evaluation. Given \textsc{CaSiRe} causal-source labels $C_i$, constructed using the annotation protocol in Appendix~\ref{app:casire_annotation}, we train with $\mathcal{L}_{\mathrm{att}}=\sum_{i\in V_{\mathrm{att}}}\mathrm{CE}(C_i,\boldsymbol{\alpha}_i)$, where $V_{\mathrm{att}}$ is the set of reply nodes with valid causal-source labels. \textbf{Sparse attribution regularizer.} To discourage ``everything caused everything,'' we add an entropy penalty: $\mathcal{L}_{\mathrm{spar}}=\sum_{i\in V_{\mathrm{att}}}H(\boldsymbol{\alpha}_i)$, which encourages low-entropy, localized attributions while being tuned to avoid overconfidence. \textbf{Inference.} We output $\arg\max_{a\in\mathcal{A}_D(i)}\alpha_{i,a}$ as the top-1 causal ancestor; top-$k$ ancestors are used for qualitative analysis.

\subsection{Potential outcomes and estimated effects}
\label{subsec:method_effects}

For each source node $i$ and intervention type $m$, C$^{3}$T estimates descendant-level potential outcomes by forcing the source intervention assignment to $T_i^{(m=1)}$ or $T_i^{(m=0)}$, while keeping other intervention components fixed. For each descendant $j\in\mathcal{D}_k(i)$, we compute the predicted negativity probabilities $\hat{p}^{\mathrm{neg}}_{j,1}=\hat{p}^{\mathrm{neg}}_j(T_i^{(m=1)})$ and $\hat{p}^{\mathrm{neg}}_{j,0}=\hat{p}^{\mathrm{neg}}_j(T_i^{(m=0)})$ using the modified source representation in the ancestor context of descendant $j$. The source-level descendant effect is $\Delta_{i,m}(k)=|\mathcal{D}_k(i)|^{-1}\sum_{j\in\mathcal{D}_k(i)}(\hat{p}^{\mathrm{neg}}_{j,1}-\hat{p}^{\mathrm{neg}}_{j,0})$, with $\Delta_{i,m}(k)=0$ when $\mathcal{D}_k(i)=\emptyset$. We estimate $\mathrm{ATE}_m(k)=\mathbb{E}_i[\Delta_{i,m}(k)]$ over source nodes in the evaluation split.

\subsection{Complexity and implementation notes}
\label{subsec:method_complexity}
Let $n$ be the number of nodes in a thread, $d$ the embedding dimension, and $D$ the ancestor window.
\textbf{Ancestor attention} requires $O(nD d)$ operations per layer, and \textbf{parent message passing} adds $O(nd)$.
We batch by threads (padding to the maximum nodes per batch) and restrict $D$ to a small constant (e.g., $D\in[8,32]$) to control runtime for deep trees.
\emph{Counterfactual prediction} reuses $\mathbf{b}_i$ and only swaps the treatment embedding, making the added compute linear in $M$ with a small constant in practice.

\section{Experimental Setup}
\label{sec:expsetup}

\subsection{Baselines}
\label{subsec:baselines}

We compare C$^{3}$T against two groups of baselines: \textbf{classic neural baselines} that use text and/or thread structure, and \textbf{prompt-only open LLM baselines} evaluated under a fixed prompting protocol. For the \textbf{non-LLM baselines}, we include a \textbf{text-only Transformer} with a pretrained \texttt{RoBERTa}/\texttt{DeBERTa} encoder for \textbf{node-level sentiment classification} and \textbf{parent--child shift prediction}. We also adapt \textbf{rumor-propagation graph models} by replacing their original veracity heads with \textbf{sentiment} and \textbf{shift} heads. This group includes \texttt{Bi-GCN}~\citep{BianAAAI2020}, \texttt{GACL}~\citep{SunWWW2022}, and temporal graph models \texttt{TGAT}/\texttt{TGN}~\citep{XuTGAT2020,RossiTGN2020}, where each thread is represented as a \textbf{reply graph} or \textbf{timestamp-ordered interaction graph}.

For attribution evaluation, graph and temporal baselines are augmented with the same bilinear ancestor-scoring head used by C$^{3}$T and trained on the same \textsc{CaSiRe} causal-source labels. These attribution variants do not use C$^{3}$T's treatment embeddings, counterfactual loss, or event-invariance penalty. Text-only baselines do not have access to ancestor structure and are therefore not evaluated on causal-source attribution.

For the \textbf{LLM baselines}, we evaluate open-weight instruction-tuned models, including \texttt{Llama~3}, \texttt{Gemma~2}, \texttt{Qwen2.5}, and \texttt{Mistral}~\citep{Llama3Meta2024,GemmaTeam2024,yang2024qwen25,JiangMistral7B2023}, as \textbf{prompt-only systems}. To reduce prompt-induced variability, we use \textbf{fixed templates}, \textbf{fixed demonstrations}, \textbf{fixed option order}, \textbf{deterministic decoding}, and the same \textbf{output schema} across all models. We evaluate three context settings: \textbf{node-only}, \textbf{node+parent}, and \textbf{node+ancestor summary}. For attribution, LLMs receive a numbered candidate-ancestor list and must return a single index; invalid, out-of-range, or unparsable outputs are mapped to \texttt{ABSTAIN} and counted as incorrect. LLM baselines are reported for sentiment and attribution only; shift prediction is evaluated for supervised neural models.

\subsection{Metrics}
\label{subsec:metrics}

We report \textbf{macro-F1} for node-level sentiment prediction $Y_i$ over the three sentiment classes and \textbf{macro-F1} for shift prediction $S_{p\to i}\in\{\texttt{down},\texttt{same},\texttt{up}\}$ over reply edges. Shift metrics are reported for supervised neural models. For \textbf{causal-source attribution}, we report \textbf{Top-1 accuracy} and \textbf{mean reciprocal rank (MRR)} over candidate ancestors within the evaluation window. For \textbf{causal effects}, we report $\mathrm{ATE}_m(k)$ for each intervention type $m$ with \textbf{95\% bootstrap confidence intervals} computed via \textbf{thread-level resampling}~\citep{Efron1979}.

\subsection{Training Details}
\label{subsec:training}

We train the main C$^{3}$T model with \textbf{\texttt{DeBERTa-base}} as the pretrained text backbone~\citep{HeDeBERTa2021}, set $d$ to the backbone hidden size, use $L\in\{2,3\}$ tree-temporal encoder layers and ancestor window $D\in\{8,16,32\}$, apply \textbf{dropout} of \texttt{0.1}, and optimize with \textbf{\texttt{AdamW}}~\citep{LoshchilovHutter2019}. We select hyperparameters on the development split, \textbf{early-stop} using development sentiment macro-F1 with patience \texttt{3}, report results over \textbf{3 random seeds}, batch examples by threads with node-level padding, and truncate each post to a fixed token budget. All non-LLM baselines use the same \textbf{preprocessing}, \textbf{event-level split}, and \textbf{evaluation metrics} wherever applicable; for graph baselines (\texttt{Bi-GCN}/\texttt{GACL}/\texttt{TGAT}/\texttt{TGN}), we follow recommended settings and tune only \textbf{learning rate} and \textbf{dropout} on the development set~\citep{BianAAAI2020,SunWWW2022,XuTGAT2020,RossiTGN2020}. For LLM baselines, we keep prompts fixed and log the \textbf{context limit}, \textbf{prompt template ID}, \textbf{few-shot} $k$, \textbf{decoding setup}, and \textbf{maximum generation length}, varying only the model checkpoint~\citep{Llama3Meta2024,GemmaTeam2024,yang2024qwen25,JiangMistral7B2023}. Full templates and parsing rules are provided in Appendix~\ref{sec:appendix_repro}.

\section{Results}
\label{sec:results}

\subsection{Main quantitative results}
\label{subsec:main_results}
Table~\ref{tab:main_results} summarizes results on \emph{node sentiment} ($Y_i$), \emph{edge-level shift} ($S_{p\rightarrow i}$), and \textbf{causal ancestor attribution} ($C_i$), under \textbf{in-domain} evaluation (held-out threads from seen events) and \textbf{event-OOD} evaluation (held-out events; primary). Across both regimes, C$^{3}$T performs best overall, with the largest gains in \emph{attribution} and \textbf{event-OOD robustness}.

\begin{table*}[t]
\centering
\small
\setlength{\tabcolsep}{3pt} 
\resizebox{\textwidth}{!}{%
\begin{tabular}{l|ccc|ccc}
\hline
& \multicolumn{3}{c|}{\textbf{In-domain}} & \multicolumn{3}{c}{\textbf{Event-OOD (held-out events)}} \\
\textbf{Model} &
\textbf{Sent.\ F1} & \textbf{Shift F1} & \textbf{Attr.\ Top-1 / MRR} &
\textbf{Sent.\ F1} & \textbf{Shift F1} & \textbf{Attr.\ Top-1 / MRR} \\
\hline
Text-only Transformer & 63.4 $\pm$ 0.7 & 48.2 $\pm$ 0.9 & \texttt{--} / \texttt{--} & 49.5 $\pm$ 1.2 & 39.1 $\pm$ 1.5 & \texttt{--} / \texttt{--} \\
Bi-GCN (adapted) \citep{BianAAAI2020} & 65.1 $\pm$ 0.5 & 55.7 $\pm$ 0.6 & 23.4 / 35.8 & 51.2 $\pm$ 1.0 & 44.3 $\pm$ 1.1 & 19.8 / 31.2 \\
GACL (adapted) \citep{SunWWW2022} & 66.8 $\pm$ 0.6 & 57.2 $\pm$ 0.5 & 26.1 / 38.5 & 55.9 $\pm$ 0.9 & 48.6 $\pm$ 0.8 & 22.4 / 34.1 \\
Temporal GNN (TGN/TGAT) \citep{RossiTGN2020,XuTGAT2020} & 66.2 $\pm$ 0.8 & 56.5 $\pm$ 0.7 & 25.8 / 37.9 & 54.1 $\pm$ 1.1 & 47.2 $\pm$ 1.3 & 21.5 / 33.7 \\
\hline
\textbf{C$^{3}$T (ours)} & \textbf{68.5} $\pm$ \textbf{0.4} & \textbf{61.3} $\pm$ \textbf{0.5} & \textbf{41.7} / \textbf{59.4} & \textbf{60.4} $\pm$ \textbf{0.6} & \textbf{54.8} $\pm$ \textbf{0.7} & \textbf{36.2} / \textbf{52.1} \\
\hline
\end{tabular}%
}
\vspace{-2mm}
\caption{\textbf{Main results.} Sentiment and shift scores are reported as \textbf{mean $\pm$ standard deviation} across 3 random seeds. Attribution is reported as seed-averaged Top-1 accuracy / MRR. C$^{3}$T demonstrates superior robustness and stability, particularly in the primary Event-OOD evaluation.}
\label{tab:main_results}
\vspace{-3mm}
\end{table*}

\subsection{Open LLM baselines}
\label{subsec:llm_results}
Table~\ref{tab:llm_results} reports \textbf{open-weight LLM prompting baselines} under the fixed protocol, comparing \emph{zero-shot (ZS)} vs.\ \emph{few-shot (FS)} and three context settings (\texttt{node-only}, \texttt{node+parent}, \texttt{node+ancestor summary} bounded by the context limit). Beyond sentiment classification, we evaluate \textbf{attribution} by prompting the LLM to select the prior message most responsible for the target sentiment.

\begin{table}[t]
\centering
\small
\setlength{\tabcolsep}{3pt} 
\resizebox{\columnwidth}{!}{%
\begin{tabular}{l|cc|cc}
\hline
\textbf{LLM / Context} &
\multicolumn{2}{c|}{\textbf{Sentiment (macro-F1)}} &
\multicolumn{2}{c}{\textbf{Attribution (Top-1 / MRR)}} \\
& \textbf{ZS} & \textbf{FS} & \textbf{ZS} & \textbf{FS} \\
\hline
Llama3 (Node) \citep{Llama3Meta2024} & 46.5 $\pm$ 1.8 & 50.8 $\pm$ 1.6 & 12.4 / 21.5 & 14.8 / 24.2 \\
Llama3 (Node+Parent) & 52.3 $\pm$ 1.5 & 55.4 $\pm$ 1.4 & 17.1 / 28.6 & 20.5 / 31.9 \\
Llama3 (Node+AncSum) & 56.8 $\pm$ 1.3 & 59.2 $\pm$ 1.2 & 21.4 / 33.7 & 25.1 / 38.4 \\
\hline
Qwen2.5 (Node+AncSum) \citep{yang2024qwen25} & 56.2 $\pm$ 1.4 & 58.9 $\pm$ 1.3 & 22.1 / 34.2 & 25.8 / 39.1 \\
Mistral (Node+AncSum) \citep{JiangMistral7B2023} & 54.1 $\pm$ 1.6 & 56.7 $\pm$ 1.5 & 19.8 / 31.5 & 23.4 / 36.8 \\
Gemma2 (Node+AncSum) \citep{GemmaTeam2024} & 55.5 $\pm$ 1.5 & 57.8 $\pm$ 1.3 & 20.6 / 32.8 & 24.2 / 37.5 \\
\hline
\end{tabular}%
}
\vspace{-2mm}
\caption{\textbf{Open LLM baselines.} ZS: zero-shot; FS: few-shot with fixed $k$ demonstrations. Sentiment scores are reported as mean macro-F1 $\pm$ half-width of the 95\% thread-level bootstrap CI. Attribution is reported as Top-1 accuracy / MRR. Context variants are Node, Node+Parent, and Node+AncSum.}

\label{tab:llm_results}
\vspace{-3mm}
\end{table}

\subsection{Generalization and social-media interpretation}
\label{subsec:generalization_interp}
C$^{3}$T reduces the \textbf{event-OOD gap} relative to in-domain performance (Table~\ref{tab:main_results}), suggesting that \emph{intervention-aware}, \emph{event-invariant} modeling yields representations less tied to event-specific surface cues (\texttt{names}, \texttt{places}, \texttt{hashtags}) and more tied to conversational function (e.g., \texttt{evidence-backed corrections} vs.\ \texttt{hostile attacks}). The largest gains are for \textbf{causal ancestor attribution}, indicating more reliable identification of earlier messages that shape affective reactions rather than defaulting to the immediate parent. Prompted LLM baselines benefit from added context (\texttt{Node+Parent}/\texttt{Node+AncSum}) but remain less stable for attribution, highlighting limitations of \emph{prompt-only} handling of long, branching threads. Overall, \textbf{structure- and intervention-aware counterfactual learning} improves sentiment-dynamics modeling beyond post-level classification.

\section{Ablations and Robustness}
\label{sec:ablations}

We ablate key components of C$^{3}$T to isolate which design choices drive event-OOD performance on (i) \emph{sentiment prediction}, (ii) \emph{shift detection}, and (iii) \textbf{causal ancestor attribution}. Unless stated otherwise, all ablations use the same \textbf{event-level split}, identical training budget, and identical hyperparameter search ranges as the full model, and results are reported on the held-out \textbf{event-OOD test set}. \textbf{Core ablations.} We run four core ablations: \textbf{A1} removes the \emph{counterfactual loss} by setting $\lambda_{\mathrm{cf}}=0$; \textbf{A2} removes \emph{intervention tags} by setting $\mathbf{u}_i=\mathbf{0}$; \textbf{A3} removes \emph{event-level predictive invariance} by setting $\lambda_{\mathrm{irm}}=0$; and \textbf{A4} restricts attribution candidates to the immediate parent only, $\mathcal{A}_D(i)=\{\pi(i)\}$, instead of using the full ancestor window. These ablations test the roles of counterfactual regularization, explicit intervention signals, event-level invariance, and ancestor-window modeling.

\begin{table}[t]
\centering
\small
\setlength{\tabcolsep}{3pt} 
\resizebox{\columnwidth}{!}{%
\begin{tabular}{l|ccc}
\hline
\textbf{Model / Ablation} & \textbf{Sent.\ F1} & \textbf{Shift F1} & \textbf{Attr.\ Top-1 / MRR} \\
\hline
C$^{3}$T (full) & \textbf{60.4} $\pm$ \textbf{0.6} & \textbf{54.8} $\pm$ \textbf{0.7} & \textbf{36.2} / \textbf{52.1} \\
\hline
A1: w/o counterfactual loss & 58.1 $\pm$ 0.8 & 52.9 $\pm$ 0.9 & 32.4 / 48.5 \\
A2: w/o intervention tags ($T_i$) & 55.7 $\pm$ 1.1 & 49.6 $\pm$ 1.2 & 25.1 / 38.9 \\
A3: w/o event invariance & 56.9 $\pm$ 0.9 & 51.3 $\pm$ 1.0 & 34.1 / 50.2 \\
A4: parent-only attribution & 59.2 $\pm$ 0.5 & 53.7 $\pm$ 0.6 & 21.5 / 21.5 \\
\hline
\end{tabular}%
}
\vspace{-2mm}
\caption{\textbf{Core ablations on the event-OOD test set.} Sentiment and shift scores are reported as \textbf{mean $\pm$ standard deviation} across 3 seeds. Attribution is reported as seed-averaged Top-1 accuracy / MRR. The sharp drop in attribution for A2 and A4 validates the necessity of intervention tags and ancestor modeling.}
\label{tab:ablations_main}
\vspace{-3mm}
\end{table}

\section{Causal Effects and Social Insights}
\label{sec:effects}
Beyond predictive performance, we use C$^{3}$T’s \emph{counterfactual branch} to derive \textbf{interpretable social findings}: how conversational interventions shape downstream sentiment trajectories and what thread-level narratives explain these patterns. All effects are \emph{observational}, and should be interpreted as \textbf{model-based causal estimates} conditional on measured context.

\subsection{Estimated effects of interventions}
\label{subsec:effects_main}
We estimate intervention effects via the \textbf{$k$-hop ATE}: for each intervention type $m$ in our taxonomy (\texttt{claim/assertion}, \texttt{denial/correction}, \texttt{evidence/link}, \texttt{authority citation}, \texttt{toxicity/attack}, \texttt{sarcasm/irony}, \texttt{question/challenge}, and \texttt{derail/off-topic}), we compute the expected change in \emph{downstream negativity} within $k$ hops when toggling $T_{i,m}$ from \texttt{0} to \texttt{1} while holding the remaining intervention components fixed, and report \textbf{95\% thread-level bootstrap CIs} \citep{Efron1979}; Table~\ref{tab:ate_main} summarizes the results.

\begin{table}[t]
\centering
\small
\setlength{\tabcolsep}{4pt}
\resizebox{\columnwidth}{!}{%
\begin{tabular}{l|c c}
\hline
\textbf{Intervention (source node)} & \multicolumn{2}{c}{\textbf{$\mathrm{ATE}_m(k)$ on downstream negativity [95\% CI]}} \\
\hline
Denial / correction & $-0.142$ & $[-0.178, -0.106]$ \\
Evidence / link     & $-0.118$ & $[-0.153, -0.083]$ \\
Authority citation  & $-0.096$ & $[-0.135, -0.058]$ \\
\hline
Question / challenge& $-0.015$ & $[-0.062, +0.032]$ \\
Claim / assertion   & $+0.032$ & $[-0.012, +0.076]$ \\
\hline
Derail / off-topic  & $+0.048$ & $[+0.005, +0.091]$ \\
Sarcasm / irony     & $+0.064$ & $[+0.015, +0.113]$ \\
Toxicity / attack   & $+0.235$ & $[+0.192, +0.278]$ \\
\hline
\end{tabular}%
}
\vspace{-2mm}
\caption{Estimated \textbf{2-hop} Average Treatment Effects ($\mathrm{ATE}_m(k)$ with $k=2$) for all $M=8$ conversational interventions on downstream negativity. Rows are sorted by effect magnitude (De-escalation $\to$ Escalation). CIs computed via thread-level bootstrap \citep{Efron1979}. Note that \emph{Question} and \emph{Claim} cross zero, indicating context-dependent effects.}
\label{tab:ate_main}
\vspace{-3mm}
\end{table}

\paragraph{Interpretation (social-media lens).}
Across events, \emph{denials/corrections} and \emph{evidence provision} tend to reduce \textbf{downstream negativity} when they occur early in a thread, consistent with the idea that clarifying uncertainty can \emph{de-escalate affect} and reduce antagonistic replies. In contrast, \emph{toxicity/attack} exhibits a \textbf{positive ATE} on negativity, indicating that hostile discourse acts as an \emph{escalation trigger} whose effect propagates beyond the immediate parent--child exchange. \emph{Questions/challenges} show \textbf{mixed effects}: in some events they function as epistemic ``speed bumps'' (reducing negativity), while in others they act as confrontational prompts (increasing negativity), suggesting that the \emph{form} of a question and the \emph{event context} jointly shape downstream tone. \emph{Sarcasm} also shows context-dependent effects, reflecting platform norms where irony may be used either to ridicule (\texttt{escalation}) or to signal in-group alignment (\texttt{stabilization}).

\subsection{Heterogeneous effects}
\label{subsec:effects_hetero}
We conduct exploratory checks for whether intervention effects vary by \textbf{information quality} and \textbf{conversational stage} when the required metadata are available. Specifically, we inspect effect patterns across veracity/stance strata and compare early versus late source nodes using timestamp quartiles and structural depth. These analyses are used only to contextualize the aggregate ATE results in Table~\ref{tab:ate_main}; we do not treat them as primary findings unless the corresponding subgroup estimates and confidence intervals are reported.

\subsection{Qualitative case studies}
\label{subsec:effects_cases}
\begin{figure}[t]
\centering
\resizebox{\columnwidth}{!}{%
\begin{tikzpicture}[
    node distance=0.8cm and 0.4cm,
    box/.style={
        rectangle,
        rounded corners=3pt,
        draw=gray!40,
        very thick,
        align=left,
        font=\sffamily\small,
        inner sep=6pt,
        text width=0.85\linewidth,
        fill=white
    },
    root/.style={box, fill=gray!5, draw=gray!60},
    toxic/.style={box, fill=red!5, draw=cbRed, line width=1.5pt},
    negative/.style={box, fill=orange!5, draw=cbOrange},
    annotation/.style={
        rectangle,
        draw=cbBlue,
        dashed,
        fill=white,
        text width=3.2cm,
        align=left,
        font=\footnotesize\bfseries,
        rounded corners
    },
    arrow/.style={-Stealth, thick, gray},
    cf_arrow/.style={-Stealth, thick, cbBlue, dashed}
]


    \node[root] (root) {
        \textbf{User A (Root):} Just heard that the new policy might actually double the tax rate? Can anyone confirm? \#Politics \\
        \textcolor{gray!80}{\tiny [Timestamp: 2:00 PM] \quad [Interventions: Question/Challenge]}
    };

    \node[toxic, below=of root] (toxic_node) {
        \textbf{User B (Ancestor):} @UserA Only a complete idiot would believe that propaganda. Do some basic research before posting garbage. \\
        \textcolor{cbRed}{\tiny [Interventions: Toxicity/Attack]}
    };

    \node[negative, below=of toxic_node, xshift=-4.4cm] (reply1) {
        \textbf{User C:} @UserB Wow, calling people idiots? Typical behavior from your side. You are the problem. \\
        \textcolor{cbOrange}{\tiny [Sentiment: Negative \quad Pred: Negative]}
    };

    \node[negative, below=of toxic_node, xshift=4.4cm] (reply2) {
        \textbf{User D:} @UserB So rude for no reason. This is why nobody takes you seriously. Blocked. \\
        \textcolor{cbOrange}{\tiny [Sentiment: Negative \quad Pred: Negative]}
    };

    \draw[arrow] (root.south) -- (toxic_node.north);
    \draw[arrow] (toxic_node.south) -- (reply1.north);
    \draw[arrow] (toxic_node.south) -- (reply2.north);


    \node[annotation, right=of toxic_node, xshift=0.2cm, yshift=0.5cm, draw=cbRed] (attrib) {
        \textcolor{cbRed}{Driver of Escalation}\\
        \normalfont\scriptsize \textit{Identified as primary cause for both User C and User D.}
    };
    \draw[dashed, cbRed, thick, ->] (attrib.west) -- (toxic_node.east);

    \node[annotation, left=of reply1, xshift=-0.2cm, draw=cbBlue] (cf1) {
        \textcolor{cbBlue}{Counterfactual}\\
        \normalfont\scriptsize If \texttt{Attack} removed:\\
        $\hat{P}(\text{Neg}) \downarrow \mathbf{0.22}$\\
        \textit{(Escalation avoided)}
    };
    \draw[cf_arrow] (cf1) -- (reply1);

    \node[above=of reply2, yshift=-0.4cm, font=\scriptsize\bfseries\color{cbBlue}] (cf2_text) {$\hat{P}(\text{Neg}) \downarrow 0.19$};

\end{tikzpicture}%
}
\caption{Case Study B (Escalation). This anonymized and paraphrased dataset-derived example illustrates a \textbf{Toxicity/Attack} intervention (User B) identified by C$^{3}$T as the primary driver of negative sentiment for multiple descendants (Users C and D). Removing the attack label reduces predicted negativity, corresponding to a positive treatment-on effect for toxicity under the ATE convention.}
\label{fig:case_study_2}
\end{figure}
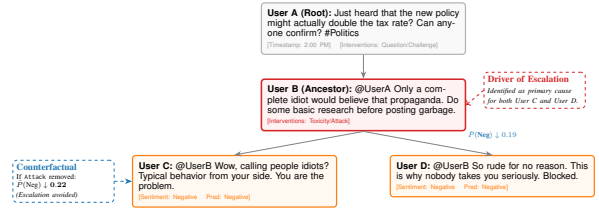

To make the \textbf{causal attributions} concrete, we present two anonymized and paraphrased case studies derived from held-out rumor-thread examples (Figure~\ref{fig:case_study_1} and Figure~\ref{fig:case_study_2}). The examples preserve the relevant conversational structure, intervention labels, sentiment labels, and model predictions, but do not reproduce original user text verbatim. Each case study includes a \emph{tree snippet}, the predicted causal ancestor for a target reply, and a \emph{counterfactual sentiment change} when removing a key intervention. For readability, the figures report the effect of removing an observed intervention, $R_{i,m}=\hat{p}^{\mathrm{neg}}(T_i^{(m=0)})-\hat{p}^{\mathrm{neg}}(T_i^{(m=1)})$, whereas Table~\ref{tab:ate_main} reports the ATE using the treatment-on minus treatment-off convention. Thus, for de-escalating interventions such as denial/correction, a positive removal effect corresponds to a negative ATE. \textbf{Case 1 (correction-driven de-escalation).} In Thread~A, an early \emph{denial/correction} is selected by C$^{3}$T as the dominant causal ancestor for a later neutral reply. Removing the denial label raises the target's predicted negativity by $R_{i,\texttt{denial}}=+0.18$, equivalently $\Delta_{i,\texttt{denial}}=-0.18$ under the treatment-on minus treatment-off convention. \textbf{Case 2 (attack-driven escalation).} In Thread~B, a \emph{toxic/attacking ancestor} receives most attribution mass for multiple negative descendants. Removing the attack label reduces predicted negativity by $R_{i,\texttt{tox}}=-0.22$, equivalently $\Delta_{i,\texttt{tox}}=+0.22$ under the treatment-on minus treatment-off convention.

\begin{figure}[t]
\centering
\resizebox{\columnwidth}{!}{%
\begin{tikzpicture}[
    node distance=0.8cm and 1.2cm,
    box/.style={
        rectangle,
        rounded corners=3pt,
        draw=gray!40,
        very thick,
        align=left,
        font=\sffamily\small,
        inner sep=6pt,
        text width=0.85\linewidth,
        fill=white
    },
    root/.style={
        box,
        fill=gray!5,
        draw=gray!60
    },
    correction/.style={
        box,
        fill=blue!5,
        draw=cbBlue,
        line width=1.5pt
    },
    target/.style={
        box,
        fill=green!5,
        draw=cbGreen
    },
    other/.style={
        box,
        text width=0.35\linewidth,
        font=\scriptsize\sffamily,
        fill=gray!2
    },
    annotation/.style={
        rectangle,
        draw=cbRed,
        dashed,
        fill=white,
        text width=3.5cm,
        align=left,
        font=\footnotesize\bfseries,
        rounded corners
    },
    arrow/.style={-Stealth, thick, gray},
    labeltext/.style={font=\scriptsize\sffamily\bfseries, color=gray!80, anchor=south west}
]


    \node[root] (root) {
        \textbf{User A (Root):} Breaking! Unconfirmed reports of active shooter at the downtown mall. Stay away! \#Alert \\
        \textcolor{gray!80}{\tiny [Timestamp: 10:00 AM] \quad [Interventions: Claim]}
    };

    \node[correction, below=of root, xshift=-0.5cm] (reply1) {
        \textbf{User B (Reply):} @UserA This is \textbf{FALSE}. Police just tweeted it was a firecracker prank. No shooter. Link attached. \\
        \textcolor{cbBlue}{\tiny [Interventions: Denial/Correction, Evidence/Link]}
    };

    \node[other, right=of reply1, xshift=-0.5cm, yshift=0.5cm] (reply2) {
        \textbf{User C:} OMG hearing sirens everywhere!
    };

    \node[target, below=of reply1, xshift=0.5cm] (target) {
        \textbf{User D (Target):} @UserB Thanks for the update. Glad it's just a prank. I was about to panic. \\
        \textcolor{cbGreen}{\tiny [Sentiment: Neutral \quad Predicted: Neutral]}
    };

    \draw[arrow] (root.south) -- (reply1.north);
    \draw[arrow] (root.south) -| (reply2.north);
    \draw[arrow] (reply1.south) -- (target.north);


    \node[annotation, left=of reply1, xshift=0.2cm, draw=cbBlue] (attrib) {
        \textcolor{cbBlue}{Primary Causal Ancestor}\\
        \normalfont\scriptsize Model assigns $P(\text{cause}) = 0.72$\\
        \textit{"Correction drives de-escalation"}
    };
    \draw[dashed, cbBlue, thick, ->] (attrib) -- (reply1);

    \node[annotation, right=of target, xshift=-0.2cm, draw=cbRed] (cf) {
        \textcolor{cbRed}{Counterfactual Effect}\\
        \normalfont\scriptsize If \texttt{Denial} removed:\\
        $\hat{P}(\text{Negativity}) \uparrow \mathbf{0.18}$\\
        \textit{(Sentiment would shift to Negative)}
    };
    \draw[dashed, cbRed, thick, ->] (cf) -- (target);

\end{tikzpicture}%
}
\caption{Case Study A (De-escalation). This anonymized and paraphrased dataset-derived example illustrates a \textbf{Denial/Correction} intervention (User B) identified by C$^{3}$T as the primary causal ancestor for the subsequent neutral reply (User D). Removing the correction tag increases predicted negativity by $R_{i,\texttt{denial}}=+0.18$, corresponding to $\Delta_{i,\texttt{denial}}=-0.18$ under the treatment-on minus treatment-off convention.}
\label{fig:case_study_1}
\end{figure}
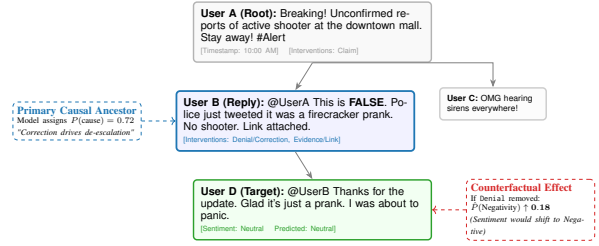

\section{Additional Analysis}
\label{sec:additional-analysis}

\paragraph{Data and structural scope.}
We evaluate on the public PHEME/RumourEval conversation-tree benchmarks. In the primary event-OOD setting, training contains 980 threads from \textit{charliehebdo}, \textit{ferguson}, and \textit{germanwings-crash}; development contains 175 \textit{ottawashooting} threads; and testing contains 205 \textit{sydneysiege} threads (Table~5). No tree is shared across partitions; full reconstructable reply structures are retained. The available statistics omit post and edge totals, mean descendants, leaf percentage, and ATE-eligible nodes; we make no claims about these properties. C$^3$T uses the nearest $D=16$ ancestors for attribution and $k=2$ descendants for intervention analysis; replies without candidates are excluded, while sources without two-hop descendants receive zero by convention. The reported effects therefore characterize short-range propagation within observed rumor-centered trees, not exposure outside the tree, cross-branch reading, or longer-range diffusion.

\paragraph{Counterfactual estimand and assumptions.}
For intervention type $m$, C$^3$T forces only $T_{i,m}$ on or off while holding source text, reply structure, temporal context, and other intervention components fixed. This yields a conditional, model-based sensitivity of predicted downstream negativity. It does not simulate rewriting or deleting a post, which could change wording, participation, replies, and structure. The potential-outcomes interpretation relies on temporal precedence, consistency, positivity, limited interference, and conditional ignorability given observed text, position, time, and history (Section~2.5). Conditional ignorability is the nonspuriousness assumption in this formulation, but unobserved beliefs, offline exposure, community membership, and platform processes may still confound the estimates. Table~4 therefore does not provide randomized real-world effects.

\paragraph{What the objective and ablations establish.}
The counterfactual loss separates treatment-on/off predictions for observed interventions and suppresses absent-intervention sensitivity. It does not encode effect direction, so the signs in Table~4 are not directly imposed, although the objective may affect magnitude. The $\lambda_{\mathrm{cf}}=0$ ablation evaluates prediction and attribution, not ATE stability: event-OOD sentiment macro-F1 falls from $60.4$ to $58.1$, shift macro-F1 from $54.8$ to $52.9$, and attribution Top-1/MRR from $36.2/52.1$ to $32.4/48.5$. Removing intervention tags yields $55.7$, $49.6$, and $25.1/38.9$, respectively; parent-only attribution yields $21.5/21.5$. These results support counterfactual regularization, explicit intervention signals, and multi-ancestor modeling, but do not establish invariance of ATE magnitudes to the loss or tags.

\paragraph{Intervention provenance and label uncertainty.}
The eight intervention types were defined from prior work; the LLM did not generate the taxonomy. Llama~3-8B only assigns the predefined multi-label tags. Scores are temperature-calibrated, thresholded by type ($0.65$ for denial, $0.70$ for evidence, $0.80$ for toxicity, and $0.50$ otherwise), and abstained below $0.4$, excluding $4.2\%$ of test nodes from the counterfactual loss. The separately reported Fleiss' $\kappa=0.75$ measures three-annotator agreement for causal-source labels, not validation of LLM intervention tags. Calibration and abstention reduce uncertainty, but errors in sarcasm, indirect hostility, coded language, or ambiguous challenges may propagate into effect estimates.

\paragraph{Interpreting the reported effects and cases.}
Within this estimand, forcing denial/correction ($-0.142$), evidence/link ($-0.118$), or authority-citation ($-0.096$) embeddings on is associated with lower predicted two-hop negativity, while forcing toxicity/attack on is associated with higher predicted negativity ($+0.235$); their Table~4 bootstrap intervals do not cross zero. Intervals for question/challenge and claim/assertion cross zero, so no stable aggregate direction is supported. The 1,000 thread-level resamples capture across-tree variation, not tagger or objective uncertainty. In the selected, anonymized case studies, removing a denial tag raises predicted negativity by $0.18$, while removing a toxicity tag lowers it by $0.22$. Because no case-pattern frequency is reported, these examples illustrate behavior, not prevalence, and are weaker evidence than the aggregate estimates.

\paragraph{Robustness boundary.}
The current results do not report effects under permuted tags, a prevalence-matched placebo, alternative $\lambda_{\mathrm{cf}}$, $\gamma$, or $\beta$, or removal of the counterfactual loss. We therefore make no claim that magnitudes or rankings are stable under these diagnostics. Together with per-type human validation of intervention tags, these checks are needed to separate data-supported sensitivity from sensitivity induced by tagging or training. Our conclusions remain limited to the reported signs, intervals, predictive ablations, and examples.

\section{Conclusion}
\label{sec:conclusion}
We introduce \textsc{CaSiRe} and C$^{3}$T for \textbf{sentiment}, \textbf{shift}, \textbf{causal-ancestor attribution}, and \textbf{counterfactual intervention reasoning} in rumor trees. Under \textbf{event-level OOD evaluation}, C$^{3}$T improves \textbf{attribution} and \textbf{generalization} by integrating \textbf{thread structure}, \textbf{temporal context}, and \textbf{intervention-aware modeling}. Identifying \textbf{who influenced whom} and \textbf{which discourse moves changed sentiment} clarifies dynamics in social media. Future work will extend the framework across \textbf{languages}, \textbf{platforms}, and \textbf{policy settings}.

\section*{Limitations}
\label{sec:limitations}

Our effects are estimated from \emph{observational social-media data} and are therefore \textbf{conditional and model-based}. Although C$^{3}$T conditions on \texttt{text}, \texttt{reply structure}, \texttt{temporal order}, and \texttt{intervention signals}, \emph{unmeasured confounding} may remain. For example, a user's prior beliefs, offline exposure to the event, community affiliation, or platform-specific norms may influence both the type of message they post and the sentiment expressed by later replies. Thus, our estimated intervention effects should not be interpreted as \textbf{randomized causal effects}; rather, they are counterfactual estimates under the observed covariates and modeling assumptions.

A second limitation concerns \textbf{interference}. We model effects primarily within bounded $k$-hop descendant neighborhoods, but social-media users may read messages without replying, respond across branches, or be influenced by parallel threads about the same event. As a result, the \emph{limited-interference assumption} is only approximate. Cross-branch spillovers, quote-post behavior, external news updates, and platform recommendation systems may all shape downstream sentiment in ways that are not fully captured by the local reply tree. Future work could incorporate \texttt{exposure logs}, \texttt{cross-thread links}, and \texttt{temporal event timelines} to better model these broader interaction effects.

The \textbf{intervention labels} are another source of uncertainty. In our setup, discourse moves such as \texttt{correction}, \texttt{evidence}, \texttt{toxicity}, \texttt{sarcasm}, and \texttt{derailment} may be produced by LLM-assisted tagging or a learned tagger trained on limited supervision. These labels are inherently noisy because many conversational moves are \emph{implicit}, \emph{context-dependent}, or \emph{culturally specific}. Sarcasm, indirect hostility, coded language, and ambiguous questions are particularly difficult to identify reliably. Although calibration, thresholding, and abstention reduce some errors, remaining label noise can propagate into attribution and counterfactual effect estimates.

Our \textbf{attribution mechanism} is also limited by the candidate ancestor window. C$^{3}$T selects causal sources from a bounded set of prior messages, which makes computation feasible and improves interpretability, but some important causes may lie outside the window or outside the observed thread altogether. In addition, sparse attribution encourages focused explanations, but real conversations may have \emph{multiple interacting causes} rather than a single dominant ancestor. Therefore, the predicted causal ancestor should be viewed as the \emph{most plausible modeled source} within the available context, not as a definitive explanation of human intent or conversational causality.

Generalization beyond \textbf{rumor-centric, event-driven conversations} is not guaranteed. The datasets used in this work are centered on public rumor and crisis-style discussions, where replies often involve uncertainty, correction, evidence, and stance-taking. Other domains, such as casual conversations, long-form forums, multilingual communities, political deliberation, or moderation-heavy platforms, may exhibit different sentiment dynamics and intervention effects. Extending the framework to \texttt{additional platforms}, \texttt{languages}, and \texttt{conversation genres} is necessary before making broader claims about online discourse.

Finally, the \textbf{counterfactual branch} operates by forcing intervention embeddings on or off while holding the non-intervention text and structure fixed. This provides a controlled way to estimate how the model associates intervention types with sentiment outcomes, but it does not generate a fully realistic alternative conversation. In reality, removing a toxic attack or adding a correction would likely change the wording, subsequent replies, user participation, and thread structure. Thus, our counterfactuals are best understood as \emph{representation-level approximations} of intervention effects, not complete simulations of how the conversation would have unfolded under a different discourse history.

\section*{Ethical considerations}
\label{sec:ethics}

We use only \textbf{publicly released benchmark conversations} and follow their distribution policies. We do not attempt to de-anonymize users, infer private attributes, link accounts across platforms, or enrich the data with external identity information. Our \textsc{CaSiRe} release provides dataset-aligned identifiers and derived labels/structures; where redistribution restrictions apply, we release only \texttt{post IDs}, \texttt{annotation metadata}, and \texttt{reconstruction scripts}. We redact or hash incidental personally identifiable information in annotation notes and do not release annotator metadata. These choices are intended to support reproducibility while minimizing privacy risk for the individuals whose posts appear in the original benchmark datasets.

Rumor-centered conversations may contain \emph{harassment}, \emph{hateful language}, \emph{threats}, \emph{graphic descriptions}, or \emph{distressing crisis-related content}. To reduce annotator risk, we use content warnings, allow skip options, rotate workload, and avoid requiring repeated exposure to highly abusive material. Annotation guidelines emphasize that labels should describe the conversational function of a post rather than judge the identity or character of the user. We also avoid reproducing harmful or abusive text in the paper unless it is necessary for explanation. The qualitative examples in Figures~\ref{fig:case_study_1} and~\ref{fig:case_study_2} are anonymized and paraphrased from held-out dataset-derived cases rather than reproduced verbatim.

When \textbf{open LLMs} are used as weak labelers for intervention tags, we treat their outputs as \emph{noisy auxiliary signals} rather than ground truth. LLMs may encode social, cultural, or platform-specific biases, and their predictions may be unreliable for sarcasm, implicit toxicity, coded language, dialectal variation, or multilingual content. To reduce these risks, we calibrate LLM confidence scores, apply type-specific thresholds, allow abstention under low confidence, and validate the labels on a human-labeled subset. We also report error analyses where feasible and caution that downstream causal estimates depend partly on the quality of these intervention labels.

The \textbf{causal attribution component} introduces additional ethical risks. A model that identifies a prior message as the likely driver of downstream negativity could be misused to target, blame, or harass individual users. To mitigate this risk, we frame the system as a \emph{research and safety-analysis tool}, not as a mechanism for assigning individual responsibility or punishment. Our primary reporting focuses on aggregate intervention effects, uncertainty estimates, and qualitative examples rather than user-level blame. Any real-world deployment would require additional safeguards, human review, appeal mechanisms, and domain-specific validation.

The broader impact of this work is \textbf{two-sided}. On the positive side, the framework may help researchers, journalists, and platform-safety teams better understand how corrections, evidence, questions, sarcasm, derailment, and toxic attacks shape conversational tone in rumor threads. Such understanding may support healthier online discussions, earlier detection of escalation, and better evaluation of corrective or moderation strategies. On the negative side, the same modeling tools could be misused to optimize manipulative messaging, suppress legitimate disagreement, or strategically influence emotional reactions in online communities. We therefore emphasize that the model should be used for \emph{diagnostic and aggregate analysis}, not for automated moderation or behavioral manipulation.

Finally, all causal estimates in this work are \textbf{observational and model-based}. They should be interpreted as conditional evidence under the observed data, model design, and stated assumptions, not as definitive proof of real-world causal effects. We explicitly acknowledge uncertainty due to \emph{unmeasured confounding}, \emph{limited interference assumptions}, \emph{label noise}, and \emph{domain shift}. We encourage future work to combine this framework with stronger validation designs, such as expert audits, cross-platform replication, prospective studies, or ethically approved field experiments where appropriate.

\bibliography{custom}

\appendix

\section{Related Work}
\label{sec:related}

\subsection{Rumor and conversation-tree modeling}
Rumor modeling on social media has evolved from \emph{text-only classification} toward models that exploit \textbf{propagation}, \textbf{reply structure}, \textbf{temporal evolution}, and \textbf{user interaction patterns}. Early neural approaches modeled rumor events as time-ordered post streams using recurrent networks \citep{MaIJCAI2016}, while later work showed that the structure of rumor diffusion provides important signals beyond local text \citep{MaACL2017,ZubiagaCOLING2016,ZubiagaPLOSONE2016}. \texttt{RumourEval} further standardized conversation-thread evaluation by introducing benchmark tasks for rumor stance and veracity over structured discussions \citep{RumourEval2017,RumourEval2019}. Subsequent models explicitly represented threads as trees or graphs, including tree-structured recursive neural networks \citep{MaACMTIST2020}, Tree-LSTM and convolutional tree models \citep{KumarCarleyACL2019}, graph-aware co-attention models \citep{LuLiGCAN2020}, user-interaction attention models \citep{KhooAAAI2020}, coupled hierarchical transformers for stance-aware verification \citep{YuACL2020}, and bidirectional graph convolutional networks such as \texttt{Bi-GCN} \citep{BianAAAI2020}. Other work incorporated user credibility, multi-task learning, and stance-aware signals to improve rumor verification \citep{LiACL2019,KochkinaACL2018}.

Recent rumor detection research has increasingly emphasized \textbf{robustness} and \textbf{temporal generalization}. Graph adversarial contrastive learning improves event-invariant rumor representations by reducing sensitivity to event-specific surface cues \citep{SunWWW2022}, while dynamic and fine-grained graph models capture evolving propagation patterns \citep{SunAAAI2022,GuoACLFindings2025}. More recent methods optimize temporal propagation structure by weighting edges with time intervals and denoising the propagation tree \citep{PengCOLING2025}, or jointly model text, tree structure, and temporal signals through temporal tree transformers \citep{WuPLOSONE2025}. A recent systematic review of graph-based rumor detection also highlights the field's shift toward integrating \texttt{content}, \texttt{propagation}, \texttt{temporal}, and \texttt{social} features \citep{AlThulaiaESWA2025}. These works establish the importance of structured and temporal modeling, but they primarily target \emph{rumor/veracity prediction}. In contrast, our work focuses on \textbf{affective dynamics}: how sentiment changes inside a reply tree, which prior message triggered the change, and how conversational interventions may causally shape downstream sentiment.

\subsection{Sentiment, emotion, and stance in conversations}
\emph{Sentiment} and \emph{emotion} are affective signals, whereas \emph{stance} captures a user's orientation toward a claim or target. Rumor conversations often contain both: a reply may support or deny a claim while also expressing fear, anger, sarcasm, or relief. Stance detection was formalized in \texttt{SemEval} \citep{MohammadSemEval2016} and later adapted to rumor-specific stance classification in tree-structured social-media conversations \citep{ZubiagaCOLING2016,RumourEval2019}. Work on rumor threads further showed that users orient to and spread rumors through conversational interaction, not only through isolated posts \citep{ZubiagaPLOSONE2016}. However, stance and veracity models usually treat affect as auxiliary or implicit, rather than modeling how affect shifts from parent to child.

Conversational emotion recognition provides useful modeling ideas for \textbf{affective dynamics}. \texttt{DialogueRNN} models speaker states and conversational context for turn-level emotion recognition \citep{MajumderDialogueRNN2019}, \texttt{MELD} provides a multimodal benchmark for multi-party emotion recognition \citep{PoriaMELD2019}, and graph-based dialogue models such as \texttt{DialogueGCN} represent speaker interactions as relational graphs \citep{GhosalDialogueGCN2019}. Later work introduced directed acyclic graph structures and context-aware memory mechanisms for emotion recognition in conversations \citep{ShenDAGERC2021,HuDialogueCRN2021}. In social-media settings, recent ICWSM work analyzes temporal sentiment and emotional shifts in comment sections \citep{HossainICWSM2024}, while newer work examines sentiment shifts across Mastodon instances during crisis-related discourse \citep{KimICWSMWorkshop2025}. Rumor-specific emotion analysis has also begun to study how emotions differ between rumor and non-rumor threads and how emotions propagate through discussion structures \citep{XingICWSMWorkshop2025}. These studies motivate modeling affect as a \emph{dynamic property of conversation}. Our work extends this line by explicitly predicting \textbf{parent--child sentiment shifts} and by attributing each shift to a likely \textbf{causal ancestor} rather than only describing aggregate sentiment trajectories.

\subsection{Causal inference in text-rich social-media settings}
Causal inference in observational social-media data is commonly framed using \textbf{potential outcomes}, \textbf{treatment assignment}, \textbf{positivity}, \textbf{consistency}, and \textbf{conditional ignorability} assumptions \citep{Rubin1974,RosenbaumRubin1983,Pearl2009,ImbensRubin2015}. In text-rich settings, language may act as a treatment, an outcome, or a confounder, making causal estimation difficult because text is high-dimensional and semantically entangled \citep{KeithACL2020,FederTACL2022}. Representation-based approaches therefore learn text embeddings that preserve information needed for causal adjustment \citep{VeitchUAI2020}, while methodological work on causal inference with text provides guidance on when language can support confounding adjustment and effect estimation \citep{EgamiSciAdv2022}. NLP work has also studied causal effects of linguistic properties and counterfactual perturbations \citep{PryzantNAACL2021}, as well as counterfactual invariance for robustness to spurious correlations \citep{VeitchNeurIPS2021}.

In online conversations, causal questions are especially challenging because replies are \emph{temporally ordered}, \emph{socially dependent}, and affected by latent user ideology, event context, and prior discourse. \citet{SridharGetoorIJCAI2019} estimate causal effects of reply tone in online debates while adjusting for latent ideological confounding, and \citet{ZhangCSCW2020} quantify causal effects of conversational tendencies in text-based counseling conversations. More recent work formalizes isolated causal effects of language, emphasizing that estimating the effect of one language-encoded intervention requires controlling for non-focal language that may otherwise induce omitted-variable bias \citep{LinICML2025}. Other recent approaches use LLMs to support text-intervention estimation, but also caution that LLMs should not be treated as reliable causal reasoners without appropriate estimation procedures \citep{GuoIJCNLPFindings2025}. Our work is closest to these \textbf{causal-text} and \textbf{causal-conversation} studies, but differs in \emph{granularity} and \emph{structure}: instead of estimating a global effect of tone or language style, we estimate \textbf{intervention-specific effects} inside full conversation trees and jointly learn node sentiment, sentiment shift, and causal-source attribution.

\subsection{Conversational interventions, attribution, and counterfactual explanations}
Understanding social-media discourse often requires identifying not only what was said, but what kind of conversational move was performed. Prior research on rumor stance, argumentation, and debate has studied support, denial, questioning, disagreement, and evidentiality as important discourse functions \citep{ZubiagaCOLING2016,RumourEval2019,SridharACL2015,TanWWW2016}. Work on misinformation and rumor correction further suggests that corrective messages, evidence, authority references, and challenges may alter subsequent discussion trajectories \citep{VosoughiScience2018,FriggeriICWSM2014,StarbirdICWSM2014}. At the same time, toxicity, insults, sarcasm, and attacks can reshape participation and escalate negative affect \citep{WulczynWWW2017,DavidsonICWSM2017,ChandrasekharanCSCW2017}. These lines motivate our intervention taxonomy, which includes claim/assertion, denial/correction, evidence/link, authority citation, toxicity/attack, sarcasm/irony, question/challenge, and derail/off-topic.

Attribution is also central to interpretable social-media analysis. Attention-based rumor and stance models identify salient users, posts, or propagation paths \citep{KhooAAAI2020,YuACL2020}, while graph-based models explain veracity prediction through structural and semantic cues \citep{LuLiGCAN2020,BianAAAI2020}. However, most attribution methods remain predictive: they identify useful evidence for a classifier rather than estimating which prior conversational act plausibly caused a downstream affective shift. Counterfactual explanations in NLP offer a complementary perspective by asking how a prediction would change under a minimally altered input \citep{KaushikICLR2020,GardnerEMNLP2020,PryzantNAACL2021}. C$^{3}$T connects these ideas by learning sparse ancestor attribution together with counterfactual treatment masking, allowing the model to ask whether removing an intervention signal such as correction or toxicity would alter predicted downstream negativity.

\subsection{LLMs for social-media analysis and weak supervision}
Instruction-tuned LLMs are increasingly used for social-media labeling, stance detection, misinformation analysis, toxicity detection, and qualitative coding because they can follow natural-language task descriptions with limited supervision \citep{JiangMistral7B2023,GemmaTeam2024,yang2024qwen25,Llama3Meta2024}. Recent work also explores LLMs for social-media annotation and platform-governance tasks, including hate-speech labeling and misinformation analysis \citep{GiorgiICWSM2025,He2024LLMSocialScience,ZiemsACL2024}. However, LLM annotations may inherit bias, vary across prompts, and produce unstable outputs under small changes in task framing or option order \citep{PezeshkpourHruschka2024,GiorgiICWSM2025,HorychNAACLFindings2025}. This motivates calibrated, controlled use of LLMs as weak labelers rather than unverified ground truth.

Our work uses LLMs in two limited roles. First, LLMs provide candidate intervention tags, which are calibrated, thresholded, and allowed to abstain under uncertainty. Second, LLMs serve as prompt-only baselines for sentiment and attribution under fixed templates, decoding settings, and context budgets. This design follows recent practice in using LLMs as annotation aids while explicitly evaluating their reliability. The results further test whether prompt-only LLMs can reason over long, branching social-media threads, or whether structure-aware models remain necessary. In this respect, our work complements recent LLM-based social-media analysis but argues that counterfactual sentiment reasoning in conversation trees requires explicit modeling of structure, time, intervention signals, and causal attribution.


\section{Reproducibility Details}
\label{sec:appendix_repro}

This section describes the implementation and evaluation choices used to reproduce the experiments. We provide details on \textbf{dataset partitioning}, \textbf{preprocessing}, \textbf{\textsc{CaSiRe} annotation}, \textbf{model hyperparameters}, \textbf{intervention tagging}, \textbf{causal-effect estimation}, and \textbf{LLM prompting baselines}. All splits are defined at the \emph{thread} or \emph{event} level so that no conversation tree is shared across train, development, and test partitions.

\subsection{Dataset Splits and Partitioning Protocols}
\label{subsec:splits}

We use three complementary split protocols. The \textbf{Event-OOD split} is the primary evaluation setting. It partitions the dataset by root event ID so that all threads associated with a held-out event are excluded from training and development. This prevents leakage through repeated event names, hashtags, locations, users, and recurring claims. In this setting, \texttt{sydneysiege} is held out exclusively for testing, \texttt{ottawashooting} is used for development, and the remaining events are used for training.

The \textbf{In-domain split} evaluates standard predictive performance when train, development, and test examples come from the same event pool. We perform a stratified \texttt{80/10/10} split at the thread level within the training-event pool. A thread never appears in more than one partition. Stratification is applied over the available node-level sentiment labels so that class imbalance is approximately preserved.

The \textbf{Cross-platform split} evaluates robustness to platform shift when platform-specific subsets are available. We train and tune on Twitter threads from \texttt{charliehebdo}, \texttt{ferguson}, \texttt{ottawashooting}, and \texttt{germanwings-crash}, and test on Reddit threads from the same event set. This split is used only for robustness analysis and is not used for final model selection.

\begin{table}[ht]
    \centering
    \scriptsize
    \setlength{\tabcolsep}{3pt}
    \renewcommand{\arraystretch}{1.05}
    \caption{Event-level partitioning for the PHEME/RumourEval dataset in the primary Event-OOD setting.}
    \label{tab:event_splits}
    \begin{tabularx}{\columnwidth}{@{}lYc@{}}
        \toprule
        \textbf{Split} & \textbf{Assigned Events} & \textbf{\# Threads} \\ 
        \midrule
        \multirow{3}{*}{\textbf{Train}} 
            & \texttt{charliehebdo} & 458 \\
            & \texttt{ferguson} & 284 \\
            & \texttt{germanwings-crash} & 238 \\ 
        \midrule
        \textbf{Dev}   
            & \texttt{ottawashooting} & 175 \\ 
        \midrule
        \textbf{Test}  
            & \texttt{sydneysiege} & 205 \\
        \bottomrule
    \end{tabularx}
\end{table}

For all protocols, we preserve the full \emph{reply-tree structure} of each thread. We remove only threads whose tree cannot be reconstructed because of missing parent references or malformed timestamps. Node-level sentiment labels are obtained from the \textsc{CaSiRe} annotation layer over the original rumor-thread nodes and are used as the supervision signal for sentiment prediction. Edge-level shift labels are deterministically induced from the sentiment labels of each parent--child pair using the ordinal order \texttt{negative} $<$ \texttt{neutral} $<$ \texttt{positive}. Thus, shift labels are not separately annotated; each child reply is labeled as an upward, downward, or unchanged sentiment shift relative to its parent.

\subsection[C3T Hyperparameters and Compute]{\texorpdfstring{C$^{3}$T}{C3T} Hyperparameters and Compute}
\label{subsec:hyperparams}

We tune C$^{3}$T on the development split and select the final configuration according to development sentiment macro-F1 under the \textbf{Event-OOD setting}. We explore the number of tree-temporal encoder layers $L \in \{2,3\}$ and the ancestor-window size $D \in \{8,16,32\}$. The final configuration used for the main experiments is shown in Table~\ref{tab:hyperparams}. Unless otherwise stated, reported predictive scores are averaged over \texttt{three random seeds} using the same partitioning protocol.

\begin{table}[ht]
    \centering
    \scriptsize
    \setlength{\tabcolsep}{3pt}
    \renewcommand{\arraystretch}{1.05}
    \caption{Final hyperparameters for C$^{3}$T used in the main Event-OOD experiments.}
    \label{tab:hyperparams}
    \begin{tabularx}{\columnwidth}{@{}lY@{}}
        \toprule
        \textbf{Parameter} & \textbf{Value} \\ 
        \midrule
        Text Backbone & \texttt{DeBERTa-base} (139M parameters) \\
        Tree Encoder Layers ($L$) & 2 \\
        Ancestor Window ($D$) & 16 \\
        Hidden Dimension ($d$) & 768 \\
        Dropout & 0.1 \\
        Optimizer & \texttt{AdamW} ($\beta_1=0.9, \beta_2=0.999$) \\
        Learning Rate & $2\times10^{-5}$ \\
        Batch Size & 32 threads (padded by node count) \\
        Max Tokens per Node & 128 \\
        $\lambda_{cf}$ (Counterfactual Loss) & 1.0 \\
        $\lambda_{irm}$ (Event-Invariance Penalty) & 0.1 \\
        Early Stopping Criterion & Development sentiment macro-F1 \\
        Patience & 3 epochs \\
        \bottomrule
    \end{tabularx}
\end{table}

All C$^{3}$T runs are trained on a single \textbf{NVIDIA A100 GPU} with \texttt{80GB} memory. One training run takes approximately \texttt{3.5 hours} on average, including validation after each epoch. Baselines use the same data splits, text preprocessing, maximum token budget, and evaluation metrics wherever applicable. For causal-effect estimates, confidence intervals are computed using \texttt{1,000} thread-level bootstrap resamples so that uncertainty reflects variation across conversation trees rather than only individual posts.

\subsection{Causal Analysis and Intervention Details}
\label{app:causal_intervention_details}

We define $M=8$ conversational intervention types, shown in Table~\ref{tab:taxonomy}. These intervention types are designed to capture common discourse moves in rumor-centered conversations, including \emph{claim assertion}, \emph{correction}, \emph{evidence provision}, \emph{authority citation}, \emph{toxic attack}, \emph{sarcasm}, \emph{questioning}, and \emph{derailment}. Since a social-media post can perform multiple functions at once, each post receives a \textbf{multi-label intervention vector} rather than a single intervention label.

\begin{table}[t]
\centering
\scriptsize
\setlength{\tabcolsep}{3pt}
\renewcommand{\arraystretch}{1.05}
\caption{Intervention taxonomy used for treatment tagging ($M=8$).}
\label{tab:taxonomy}
\begin{tabularx}{\columnwidth}{@{}>{\raggedright\arraybackslash}p{0.32\columnwidth}Y@{}}
\toprule
\textbf{Intervention Type} & \textbf{Definition} \\
\midrule
Claim / Assertion & States a new unverified claim or rumor. \\
Denial / Correction & Rejects a prior claim or provides a correction. \\
Evidence / Link & Provides a URL, image, quote, screenshot, or cited source as evidence. \\
Authority Citation & Appeals to an official, institutional, journalistic, or otherwise authoritative source. \\
Toxicity / Attack & Contains insults, slurs, threats, harassment, or ad hominem attacks. \\
Sarcasm / Irony & Uses sarcasm or irony to mock a claim, user, group, or interpretation. \\
Question / Challenge & Asks for clarification, requests evidence, or challenges credibility. \\
Derail / Off-topic & Redirects the discussion to an unrelated issue or distracts from the original claim. \\
\bottomrule
\end{tabularx}
\end{table}

For \textbf{LLM-assisted intervention tagging}, we use \texttt{Llama~3-8B} to assign a confidence score $s_{i,m}$ for each intervention type $m$ at node $i$. Scores are calibrated on the development set using temperature scaling \citep{GuoCalibration2017}. After calibration, intervention labels are assigned using type-specific thresholds tuned on the development set for F1. The final thresholds are $\tau_{\texttt{denial}}=0.65$, $\tau_{\texttt{evidence}}=0.70$, $\tau_{\texttt{toxicity}}=0.80$, and $\tau=0.50$ for the remaining intervention types. We use a conservative abstention rule: if $\max_m s_{i,m}<0.4$, the intervention vector is marked as unknown and excluded from the counterfactual loss. This affects \texttt{4.2\%} of test nodes.

For \textbf{causal-effect estimation}, we use the two-hop descendant neighborhood ($k=2$) as the primary window. This captures short-range downstream propagation while avoiding overly broad attribution of event-level sentiment changes to a single source post. If a source node has no $k$-hop descendants, its downstream negativity value is set to zero by convention. Intervention effects are estimated by toggling one intervention type at a time while keeping all other observed node, structure, and intervention features fixed. These estimates should therefore be interpreted as \emph{conditional, model-based counterfactual effects} rather than randomized causal effects.

Gold shift labels are induced from annotated node-level sentiment labels. During training, C$^{3}$T predicts shifts through a separate shift head using paired parent--child representations. During evaluation, shift macro-F1 is computed against the induced gold shift labels. This avoids defining the shift evaluation target from the model's own predicted sentiments.

\subsection{\textsc{CaSiRe} Annotation Protocol}
\label{app:casire_annotation}

\textsc{CaSiRe} contains four annotation components: \textbf{post-level sentiment labels}, \textbf{induced parent--child shift labels}, \textbf{multi-label intervention tags}, and \textbf{causal-source labels}. Post-level sentiment labels are assigned using the three-class taxonomy \{\texttt{negative}, \texttt{neutral}, \texttt{positive}\}. Parent--child sentiment-shift labels are induced automatically from the sentiment labels of each reply and its parent. Multi-label intervention tags are obtained using the \textbf{LLM-assisted tagging protocol} in Section~\ref{app:causal_intervention_details}. Causal-source labels are annotated separately for the \textbf{attribution task}.

For \textbf{causal-source annotation}, annotators are shown the target post, its immediate parent, and the bounded candidate ancestor list used by the model. Annotators select the \textbf{single prior message} that most plausibly shaped the target reply's sentiment, or mark the instance as \texttt{unclear} when no source can be reliably identified from the available context. Instances marked as \texttt{unclear} are excluded from the \textbf{supervised attribution loss} and from \textbf{attribution evaluation}.

Each causal-source instance is labeled independently by \textbf{three annotators}. Disagreements are resolved through adjudication by a \textbf{senior annotator}. We measure inter-annotator agreement over all triple-annotated instances using \textbf{Fleiss' $\kappa$}, obtaining $\kappa=0.75$. Annotation guidelines instruct annotators to select the most plausible \textbf{conversational source} of the target's expressed sentiment, not the user responsible for the broader event or thread.

\subsection{LLM Prompting Baselines}
\label{app:llm_prompting}

We evaluate open-weight LLMs as \emph{prompt-only baselines} under a strict deterministic protocol. For all LLM baselines, decoding uses \texttt{temperature=0}, \texttt{top\_p=1.0}, and \texttt{max\_new\_tokens=128}. We impose a \texttt{4,096-token} context cap. When the full ancestor context exceeds this budget, we truncate from the oldest messages first so that the most recent context is preserved. This rule is fixed across all LLMs and all runs.

We evaluate three context variants. \texttt{Node} provides only the target post. \texttt{Node+Parent} provides the target post and its immediate parent. \texttt{Node+AncSum} provides the target post, the parent, and a bounded ancestor context. For \texttt{Node+AncSum}, we use the nearest $D=16$ ancestors, sort them chronologically for sentiment prediction, and format them using consistent role tags such as \texttt{[Source]} and \texttt{[Reply]}. For attribution, the same candidates are listed in reverse chronological order, so index \texttt{1} corresponds to the parent, index \texttt{2} to the grandparent, and so on. Any out-of-range index, malformed JSON, missing label, or non-integer attribution output is mapped to \texttt{ABSTAIN} and counted as incorrect.

For few-shot prompting, we prepend a fixed set of $k=8$ demonstrations sampled once from the training split using a fixed seed. The same demonstrations, ordering, label taxonomy, and output schema are reused across all LLMs and all evaluation examples. This avoids prompt-selection variability and ensures that differences across LLM baselines reflect model checkpoints and context conditions rather than different demonstrations.

\begin{promptbox}[label={lst:prompt}]{Zero-shot prompt for sentiment prediction and causal-ancestor attribution}
\RaggedRight

\pkw{SYSTEM:} You are an expert social media analyst specializing in rumor dynamics. Analyze a conversation thread.

\vspace{1mm}
\pkw{INPUT:}

\hspace{1em}\pfld{Target Post:} [TARGET\_TEXT]

\hspace{1em}\pfld{Parent Post:} [PARENT\_TEXT]

\hspace{1em}\pfld{Context:} [SUMMARY\_OR\_LIST\_OF\_TOP\_16\_ANCESTORS]

\vspace{1mm}
\pkw{TASK 1 -- SENTIMENT:}

\hspace{1em}Choose exactly one label from \{Negative, Neutral, Positive\}.

\vspace{1mm}
\pkw{TASK 2 -- ATTRIBUTION:}

\hspace{1em}Pick the single prior post that most likely caused or shaped the target post's sentiment.

\vspace{1mm}
\pfld{Candidates:}

\hspace{1em}1) [Ancestor 1: Parent]

\hspace{1em}2) [Ancestor 2: Grandparent]

\hspace{1em}\ldots

\hspace{1em}16) [Ancestor 16]

\vspace{1mm}
\pkw{OUTPUT:}

\hspace{1em}Return strict JSON only:

\hspace{1em}\pjson{\{"sentiment":"LABEL",\allowbreak "attribution\_index":INTEGER\}}

\vspace{1mm}
\hspace{1em}If the answer is uncertain or the output cannot be produced in the required schema, return:

\hspace{1em}\pwarn{ABSTAIN}

\end{promptbox}

\begin{promptbox}[label={lst:demos}]{Few-shot demonstrations used for all LLM baselines ($k=8$)}
\RaggedRight

\fewshot{Ex1}{
\pfld{Target:} ``This is absolutely fake news.''\\
\pfld{Parent:} ``Unconfirmed reports of shooter at mall.''\\
\pfld{Ctx:} [Source] ``Breaking: reports of an active shooter at the mall.''
}{
\pjson{\{"sentiment":"Negative",\allowbreak "attribution\_index":1\}}
}{}{}

\fewshot{Ex2}{
\pfld{Target:} ``Thanks for the link, that clarifies it.''\\
\pfld{Parent:} ``Police tweeted it was a prank. URL.''\\
\pfld{Ctx:} [Source] ``People are saying there was a shooting downtown.''\\
\hspace*{2.4em}[Reply] ``Police say there is no active shooter.''
}{
\pjson{\{"sentiment":"Positive",\allowbreak "attribution\_index":1\}}
}{}{}

\fewshot{Ex3}{
\pfld{Target:} ``I am relieved it was not as serious as people first said.''\\
\pfld{Parent:} ``Officials confirm the loud noise came from fireworks, not gunfire.''\\
\pfld{Ctx:} [Source] ``Several users report hearing gunshots near the station.''\\
\hspace*{2.4em}[Reply] ``Still waiting for confirmation.''\\
\hspace*{2.4em}[Reply] ``Officials confirm the loud noise came from fireworks, not gunfire.''
}{
\pjson{\{"sentiment":"Positive",\allowbreak "attribution\_index":1\}}
}{}{}

\fewshot{Ex4}{
\pfld{Target:} ``Stop spreading nonsense without proof.''\\
\pfld{Parent:} ``A friend told me the suspect escaped.''\\
\pfld{Ctx:} [Source] ``Police are responding to an incident near campus.''\\
\hspace*{2.4em}[Reply] ``A friend told me the suspect escaped.''
}{
\pjson{\{"sentiment":"Negative",\allowbreak "attribution\_index":1\}}
}{}{}

\fewshot{Ex5}{
\pfld{Target:} ``Do you have any source for that claim?''\\
\pfld{Parent:} ``They are hiding the real number of victims.''\\
\pfld{Ctx:} [Source] ``News reports mention several injuries after the incident.''\\
\hspace*{2.4em}[Reply] ``They are hiding the real number of victims.''
}{
\pjson{\{"sentiment":"Neutral",\allowbreak "attribution\_index":1\}}
}{}{}

\fewshot{Ex6}{
\pfld{Target:} ``Calling people idiots is not helping anyone.''\\
\pfld{Parent:} ``Only an idiot would believe the official statement.''\\
\pfld{Ctx:} [Source] ``Officials released a statement denying the rumor.''\\
\hspace*{2.4em}[Reply] ``Only an idiot would believe the official statement.''
}{
\pjson{\{"sentiment":"Negative",\allowbreak "attribution\_index":1\}}
}{}{}

\fewshot{Ex7}{
\pfld{Target:} ``This update makes the situation sound less alarming.''\\
\pfld{Parent:} ``Local news says the road is closed only as a precaution.''\\
\pfld{Ctx:} [Source] ``Reports say the whole area has been evacuated.''\\
\hspace*{2.4em}[Reply] ``Local news says the road is closed only as a precaution.''
}{
\pjson{\{"sentiment":"Positive",\allowbreak "attribution\_index":1\}}
}{}{}

\fewshot{Ex8}{
\pfld{Target:} ``Why would they lie about this?''\\
\pfld{Parent:} ``No evidence was found at the scene.''\\
\pfld{Ctx:} [Source] ``Some users claim the authorities are covering it up.''\\
\hspace*{2.4em}[Reply] ``No evidence was found at the scene.''
}{
\pjson{\{"sentiment":"Negative",\allowbreak "attribution\_index":2\}}
}{}{}

\end{promptbox}
\end{document}